# Consensus in the Parliament of AI: Harmonized Multi-Region CT-Radiomics and Foundation-Model Signatures for Multicentre NSCLC Risk Stratification

Shruti Atul Mali[1], Zohaib Salahuddin[1], Danial Khan[1], Yumeng Zhang[1], Henry C. Woodruff [1,2], Eduardo Ibor-Crespo[3], Ana Jimenez-Pastor[3], Luis Marti-Bonmati[4,5], Gloria Ribas[4], Silvia Flor-Arnal[4], Marta Zerunian[6], Damiano Caruso[6], Christophe Aubé[7,8], Florence Longueville[9], Caroline Caramella[10], Philippe Lambin[1,2]

[1]*Department of Precision Medicine, GROW - Research Institute for Oncology and Reproduction, Maastricht University, 6220 MD Maastricht, The Netherlands*

[2] *Department of Radiology and Nuclear Medicine, GROW - Research Institute for Oncology and Reproduction, Maastricht University, Medical Center+, 6229 HX Maastricht, The Netherlands*

[3] *Research & Frontiers in AI Department, Quantitative Imaging Biomarkers in Medicine, Quibim SL, Valencia, Spain*

[4] *Biomedical Imaging Research Group, La Fe Health Research Institute, Valencia, Spain*

[5] *Medical Imaging Department, La Fe University and Polytechnic Hospital, Valencia, Spain*

[6] *Radiology Unit, Department of Surgical and Medical Sciences and Translational Medicine, Sapienza University of Rome, Sant'Andrea Hospital, 00189 Rome, Italy*

[7] *Laboratoire HIFIH, Université d'Angers, SFR ICAT 4208, Angers 49000, France*

[8] *Department of Radiology, CHU Angers, Angers 49000, France*

[9] *Department of Medical Imaging, Nîmes University Hospital, 30029 Nîmes, France*

[10] *Department of Medical Imaging, Saint Joseph Hospital, Paris, France*

# Abstract

## Purpose

To evaluate the impact of harmonization and multi-region image feature integration on survival prediction in non-small cell lung cancer (NSCLC) patients. We assess the prognostic utility of handcrafted radiomics and pretrained foundation model (FM) deep features extracted from thoracic CT images across multiple regions, in combination with clinical data, using a multicentre dataset.

## Methods

Survival models were developed using handcrafted radiomic and FM deep features extracted from whole lung, tumor, mediastinal nodes, coronary arteries, and coronary artery calcium (CAC) scores in 876 lung cancer patients (balanced, 604 training and 272 test) from five centres. CT features were harmonized using ComBat, reconstruction kernel normalization (RKN), and RKN+ComBat. Models were constructed at the region of interest (ROI) level, in clinical + ROI combinations, and through ensemble strategies. Regularized Cox proportional hazards models were used to estimate overall survival, with performance assessed via concordance index (C-index), 5-year time-dependent area under the curve (t-AUC), and hazards ratios. SHAP (SHapley Additive exPlanations) values were used to interpret feature contributions, and consensus analysis was performed by thresholding predicted survival probabilities at fixed time horizons, retaining only patients where all best-performing ROI models agreed on the binary risk classification.

## Results

As expected, the TNM staging demonstrated some prognostic value (C-index = 0.67; hazard ratio = 2.70; t-AUC = 0.85) for the test set. The clinical + tumor texture radiomics model, with ComBat, achieved a high individual performance (C-index = 0.76; t-AUC = 0.88). FM deep features from cube size 50 voxels also showed strong predictive value when combined with clinical data (C-index = 0.76; t-AUC = 0.89). An ensemble model combining tumor, whole lung, mediastinal node, CAC, and FM features achieved a C-index of 0.71 and t-AUC of 0.79. Consensus analysis across the best-performing ROI models identified a high-confidence subset of patients with full model agreement. The consensus model achieved a 5-year t-AUC of 0.92, sensitivity of 96.8%, and specificity of 70.0%, covering 79% of valid cases.

## Conclusion

Harmonization and multi-region feature integration significantly improve survival prediction in NSCLC patients using CT imaging. Our results indicate that added benefit from multiple harmonization steps while also leveraging pretrained foundation models. The integration of interpretable radiomics, FM-derived features, and consensus modelling from different methods offers a robust and scalable approach to individualized risk stratification, especially in multicentre settings.

# 1. Introduction

Lung cancer is one of the most commonly diagnosed cancers worldwide, and is also the leading cause of malignancy-related mortality, causing about one in five cancer deaths [1]. Non-small cell lung cancer (NSCLC) is the most common type of lung cancer, and it has been identified to have low survival rates after late diagnosis, combined with limited treatment modalities [2]. Most common interventions for NSCLC treatment include surgery, chemotherapy, radiation therapy, targeted therapies, and immunotherapies tailored to molecular profiles, and combinations of the above [3]. The prognosis of NSCLC patients relies on developing a thorough treatment and management strategy of patient care. The TNM staging system is a widely accepted standard for assessing prognosis and treatment decision-making in NSCLC, where cases are stratified based on tumor size ('T'), involvement of lymph nodes ('N'), and distant metastasis ('M'). However, this system only provides a generalised prognosis system with no personalisation, which is mainly dependent on the characteristics of the tumours and/or nodes and/or metastases. It also fails to recognise other important prognostic variables, including the age and histological type of patients, which may have a strong impact [4]. With these limitations, there is an urgent need to incorporate more variables to get a more comprehensive and more customised prognosis.

In order to meet the requirement of more personalized prognostication, there is increased interest in combining "omics" data with clinical data. Radiomics has become a potential method to derive quantitative imaging biomarkers with imaging modalities, including computed tomography (CT), magnetic resonance imaging (MRI), and positron computed tomography (PET) [5]. These biomarkers are made up of radiomic features, handcrafted or derived from deep learning models, that can provide insights into tumor phenotype and spatial heterogeneity, and have demonstrated potential for predicting outcomes and supporting clinical decisions in NSCLC [6,7]. While much of the initial work focused on the primary tumour itself, several anatomically distinct regions of interest (ROIs) have been investigated in the context of lung cancer prognosis. The entire lung captures diffuse parenchymal changes that may be associated with comorbidities [8,9]. The primary tumor phenotype is crucial for NSCLC survival prediction, with its shape and texture features tightly linked to tumor aggressiveness and survival [4,6]. Mediastinal lymph nodes, on the other hand, are also critical factors for NSCLC, playing a crucial role in TNM staging, and being a key prognostic factor leading to a more advanced disease state and a poorer prognosis [10,11]. Cardiovascular imaging biomarkers obtained from PET-CT or CT scans, such as coronary artery calcification (CAC), have been linked to major adverse cardiovascular events (MACE) and poorer overall survival in NSCLC patients [12,13]. Specifically, CAC, a quantitative measure of atherosclerotic plaque burden typically assessed via dedicated non-contrast CT, has shown such associations [14]. A higher CAC score, often quantified using the Agatston method, reflects the extent of coronary artery disease and has also been linked to increased lung cancer mortality [14]. Similarly, texture features extracted from the whole lung [15] and mediastinal lymph nodes [16,17] have been associated with prognosis in prior radiomics studies. However, these anatomical regions have to date been largely investigated in isolation, and there is limited evidence comparing their combined prognostic utility within the same multi-institutional cohort.

A key limitation of radiomics-based models is the reproducibility and generalizability of models, particularly when applied across diverse multi-institutional datasets that involve varying imaging protocols, scanner types, and reconstruction parameters [18–20]. To address these issues, harmonization strategies are broadly categorized into image domain and feature domain. Image-domain approaches include methods such as histogram matching [21], neural style transfer [22], and generative adversarial-based image translation [23], which aim to standardize images but require large datasets, are susceptible to training instability, and may introduce artifacts [19]. An alternative image-domain method, reconstruction kernel normalization (RKN) [24,25], addresses variability introduced through different CT reconstruction kernels by dividing each scan into multiple frequency bands, and the energy in each frequency band is iteratively scaled to a chosen kernel-specific template. Conversely, feature domain methods such as ComBat [24] work directly on the extracted features, where the feature distributions are statistically corrected for scanner-induced batch effects.

Recent progress in deep learning (DL) has made data-driven feature extraction possible, one that can capture complex image representations outperforming handcrafted features [26,27]. One of the most recent developments in medical image analysis is the development of foundation models (FM) trained on large, sparsely labelled medical imaging datasets that can provide robust and transferable features that can be applied in various clinical tasks [28–30]. Unlike traditional supervised models, FMs are typically trained using self-supervised or unsupervised learning strategies, enabling them to learn rich, task-agnostic features from vast amounts of unannotated data. Such a strategy enables FMs to be effectively tailored to other downstream tasks, which can be highly generalized. In a recent study, Pai et al. [31] studied FM features on the LUNG1 [6] cohort for prognostic modeling in NSCLC. Notably, a simple linear classifier had the best performance of all the baselines that were tested, with the area under the receiver operating characteristic curve (AUC) of 0.638 and showed a significant risk stratification ($p<0.001$), highlighting the possibility that FMs may serve as powerful, annotation-friendly prognostic instruments with potential for broader clinical scope [31]. Nevertheless, such deep learning models are vulnerable to overfitting and can be affected by scanner-specific biases, raising concerns regarding their application in a real-world multicentric environment [32].

In this study, we provide a comprehensive benchmark of the prognostic value of radiomic features obtained from different anatomic regions of chest CT images of patients with NSCLC. We evaluate handcrafted radiomic features extracted from the whole lung, lung tumor, mediastinal lymph nodes, and coronary arteries (including coronary artery calcium score), as well as deep semantic features extracted from tumor patches using a pretrained FM for survival analysis of NSCLC patients. These regions (whole lung, tumor, mediastinal lymph nodes, and coronary arteries including CAC) and feature types (handcrafted radiomic and deep semantic from FM) were specifically chosen to provide a comprehensive and multi-faceted view of tumor characteristics, disease spread, systemic impacts, and relevant comorbidities for robust individualized risk stratification. Each region has previously demonstrated prognostic relevance in isolation, but its comparative utility and potential complementarity within the same multicentre cohort remain underexplored. To address this gap, we assess individual and combined ROI performance, both with and without integration of clinical variables, in predicting survival outcomes. In order to examine the effect of scanner variability, we examine how two harmonisation methods, RKN at the image-level and ComBat at the feature-level, affect model performance. Moreover, we use SHAP (SHapley Additive exPlanations) [33] to interpret model predictions and identify region-specific contributions to patient risk stratification. This integrated framework allows a comprehensive evaluation of radiomic and deep features in different ROIs and gives an understanding of the efficacy of harmonization methods in multicentre survival prediction. The schematic of the workflow is shown in Figure 1.

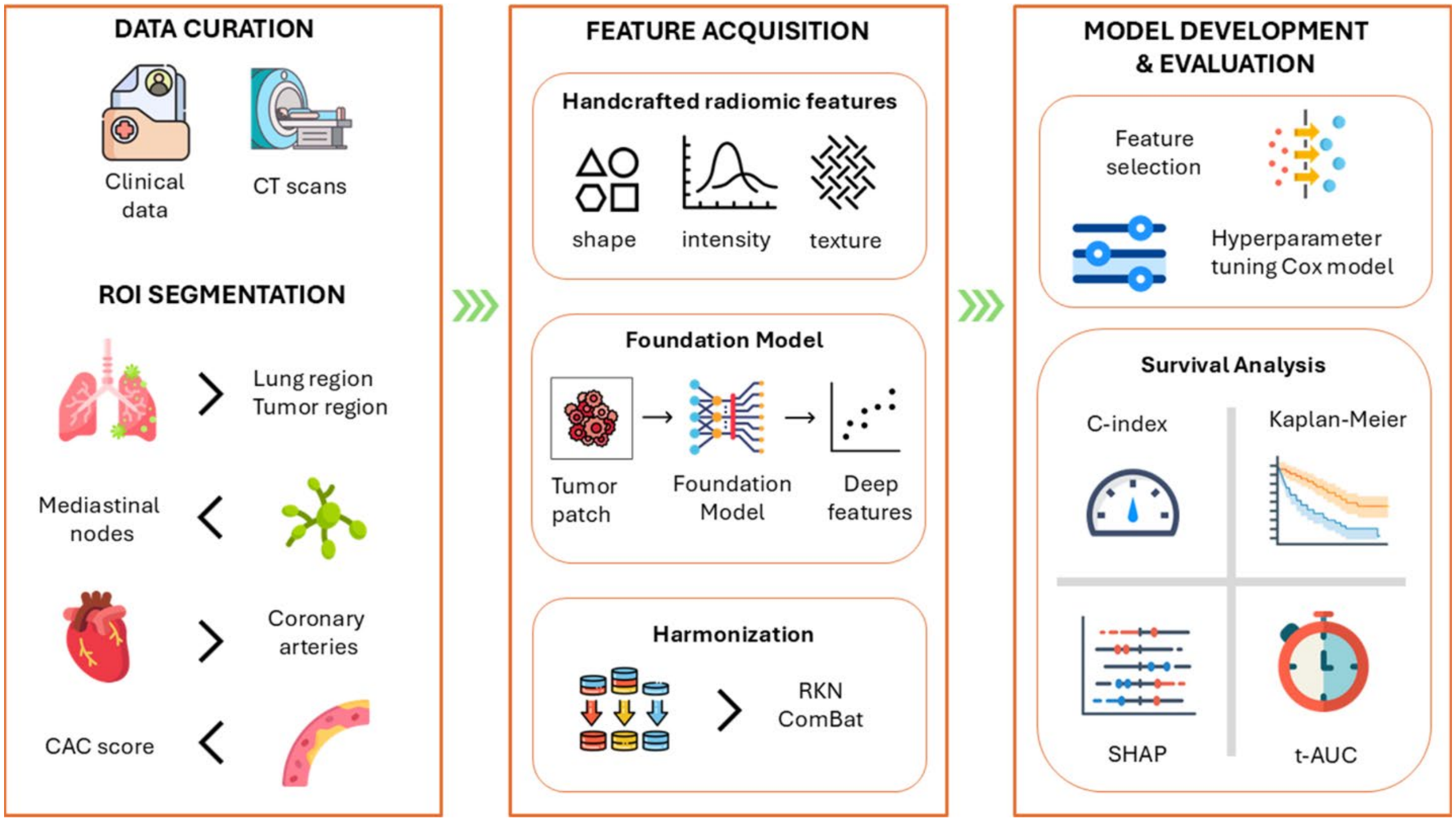

Figure 1: Overview of the survival modeling pipeline. The workflow consists of three stages: (1) Data curation and region of interest (ROI) segmentation, where clinical data and thoracic CT (computed tomography) scans are curated and segmented into the whole lung region, tumor, mediastinal nodes, coronary arteries, and coronary artery calcium (CAC) score; (2) Feature extraction, where handcrafted radiomic features (shape, intensity, texture) and foundation model (FM) deep features are extracted, followed by harmonization using Reconstruction kernel normalization (RKN) and ComBat to correct for inter-centre variability; and (3) Model development and evaluation, including feature selection, hyperparameter tuning of the Cox model, and survival analysis using concordance index (C-index), Kaplan-Meier estimation, time-dependent area under the curve (t-AUC), and SHapley Additive exPlanations (SHAP) for interpretability.

# 2. Methods

## 2.1. Data

This study utilized anonymized thoracic CT scans from the European CHAIMELEON project, a large-scale imaging repository designed to foster AI development in cancer imaging. Although CHAIMELEON hosts datasets for several cancer types (lung, breast, prostate, and colorectal cancers) [34,35], this work focuses specifically on the lung cancer cohort. Data access and model development were done within the CHAIMELEON platform, a secure, centralized infrastructure that allowed model training and evaluation while restricting raw data download. No imaging or clinical data were transferred or exported outside the platform, while survival model training was performed on the platform.

A total of 912 patients with confirmed NSCLC and baseline, pre-treatment CT scans were available, with 633 patients in the training set and 279 patients in the test set. All NSCLC patients were identified by clinicians. To ensure consistency in image resolution, the median voxel spacing across the train set (0.69, 0.69, 1 $mm^3$) was used as a reference, and patient scans with voxel spacings exceeding *mean+2* (standard deviation)* were excluded, leaving 876 patients (604 train, 272 test) for all subsequent analyses.

### 2.1.1. Study Population

The final cohort consisted of a total of 876 patients, split into 604 patients in the training set and 272 in the test set. Inclusion criteria were: (1) confirmed diagnosis of lung cancer; (2) available pretreatment CT scans; and (3) accompanying clinical and outcome data. Clinical variables included in the analysis were: age, gender, ECOG performance status, smoking status, packs/year, PD-L1 expression (in %), and TNM clinical stage. In addition, metastasis status in specific organs (brain, bone, adrenal gland, etc.) was included. Missing clinical values were imputed where necessary using appropriate strategies to ensure dataset completeness. Specifically, missing numerical values were imputed using the mean, while missing categorical values were filled with the mode. Descriptive statistics for each variable and their distributions across training and test sets are summarized in Table 1 (refer to the Results section). All statistical comparisons between the train-test sets were performed using appropriate tests based on variable type and distribution. For continuous numerical variables that were normally distributed, an Independent t-test was utilized. If continuous numerical variables were not normally distributed, the Mann-Whitney U test was employed. For categorical variables, the Chi-squared test was used to assess significant differences between the train and test sets.

### 2.1.2. Imaging Acquisition

Scans originated from five European centres (LaFe: Hospital Universitari i Politècnic La Fe (Spain), ULS: Radiology Unit at Sapienza University of Rome (Italy), CHU Angers: Centre Hospitalier Universitaire d'Angers (France), CHU Nîmes: Centre Hospitalier Universitaire de Nîmes (France), Paris St-Joseph: L'Hôpital Paris Saint-Joseph (France)) and five vendors (GE, Siemens, Philips, Toshiba, Agfa). The majority of the scans were acquired at 120 kVp, but with variability in pixel spacing and slice thickness within the datasets. Figure 2A-B illustrates patient distribution by centre and manufacturer; refer to Table 1 for the summary of image acquisition parameters.

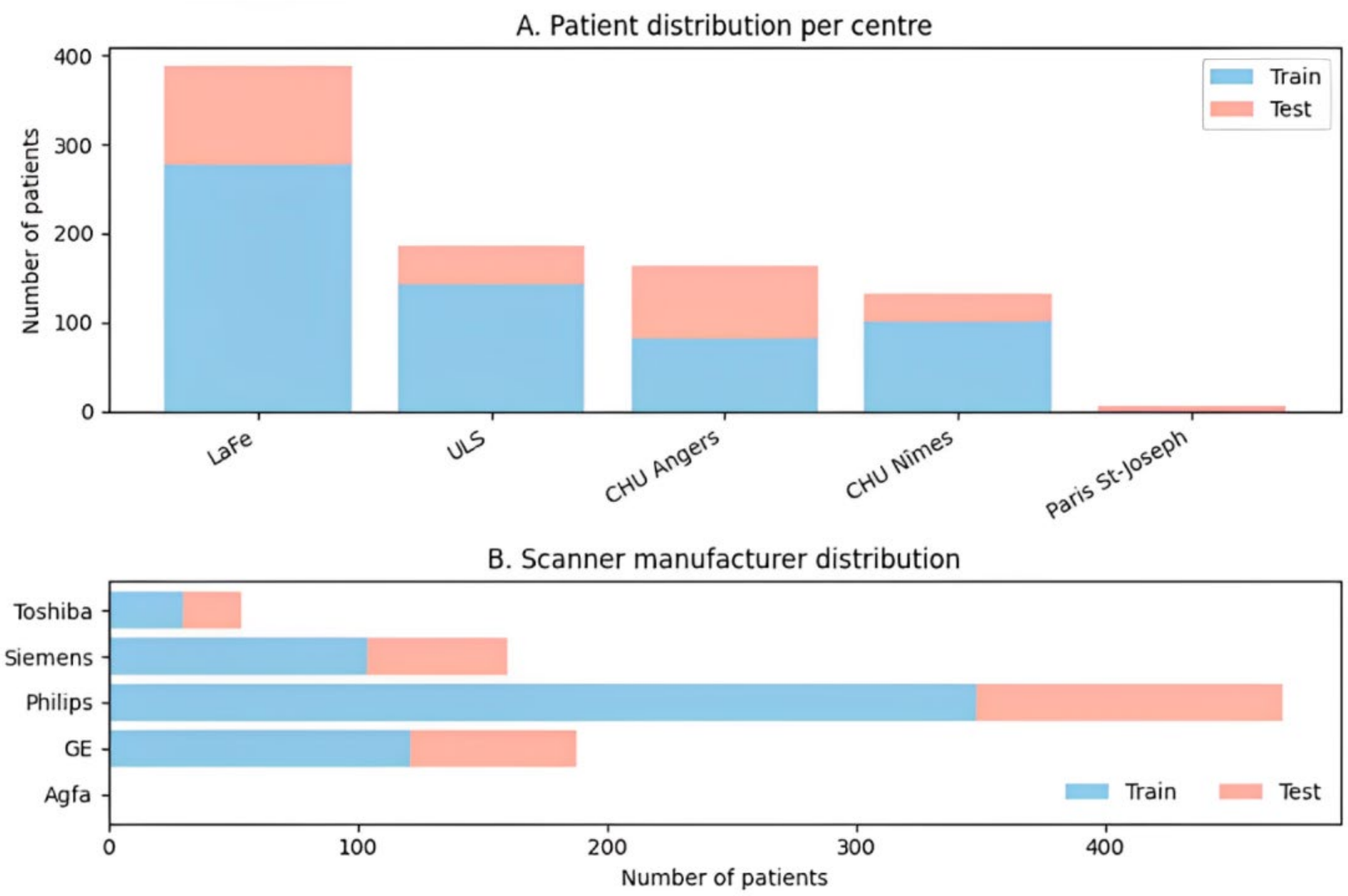

Figure 2. Patient and scanner distribution across centers.
(A) Patient distribution per acquisition centre in the training and test sets.
(B) Distribution of scanner manufacturers contributing CT scans to the dataset. Centers include LaFe (Spain), ULS (Italy), CHU Angers, CHU Nîmes, and Paris St-Joseph (France). Scanner vendors include Agfa, GE, Philips, Siemens, and Toshiba. Color bars denote a split into training (blue) and test (red) subsets.

Table 1: Imaging acquisition characteristics across training and test sets. Reported parameters include scanner manufacturer, acquisition centre, tube voltage (kVp), pixel spacing, and slice thickness. SD: standard deviation, kVp: Kilovolt Peak, LaFe: Hospital Universitari i Politècnic La Fe (Spain), ULS: Radiology Unit at Sapienza University of Rome (Italy), CHU Angers: Centre Hospitalier Universitaire d'Angers (France), CHU Nîmes: Centre Hospitalier Universitaire de Nîmes (France), Paris St-Joseph: L'Hôpital Paris Saint-Joseph (France)

| Parameter | Train (n=604) | Test (n=272) | p-value |
|---|---|---|---|
| **Manufacturer** | | | |
| Agfa | 1 (0.2%) | 0 (0%) | 0.0249 |
| GE | 121 (20%) | 67 (24.6%) | |
| Philips | 348 (57.6%) | 123 (45.2%) | |
| Siemens | 104 (17.2%) | 56 (20.6%) | |
| TOSHIBA | 30 (5%) | 23 (8.5%) | |
| **Centre** | | | |
| 01 : LaFe | 278 (46%) | 111 (40.8%) | 0.0000 |
| 03 : ULS | 142 (23.5%) | 44 (16.2%) | |
| 06 : CHU Angers | 82 (13.6%) | 81 (29.8%) | |
| 08 : CHU Nîmes | 102 (16.9%) | 30 (11%) | |
| 09: Paris St-Joseph | 0 (0%) | 6 (2.2%) | |
| | | | |
| **kVp** | | | |
| 80 | 3 (0.5%) | 0 (0%) | 0.1658 |
| 90 | 1 (0.2%) | 4 (1.5%) | |
| 100 | 142 (23.5%) | 64 (23.5) | |
| 110 | 2 (0.3%) | 3 (1.1%) | |
| 120 | 449 (74.3%) | 196 (72.1%) | |
| 130 | 5 (0.8%) | 1 (0.4%) | |
| 140 | 1 (0.2%) | 1 (0.4%) | |
| 150 | 1 (0.2%) | 0 (0%) | |
| NaN | 0 (0.0%) | 3 (1.1%) | |
| **Pixel Spacing** | | | |
| Mean (SD) | 0.67 (0.15) | 0.67 (0.17) | 0.9689 |
| Median | 0.69 | 0.70 | |
| **Slice thickness** | | | |
| Mean (SD) | 1.5 (0.89) | 1.48 (1.02) | 0.7813 |
| Median | 1.0 | 1.0 | |

## 2.2. Segmentation

### 2.2.1. Lung and lung tumor segmentation

For segmenting the lung region and lung tumors from chest CT scans, we utilized an open-source pretrained nnU-Net [36] model developed by Murugesan et al. [37–39], where the model was trained on datasets including DICOM-LIDC-IDRI-Nodules [40], NSCLC Radiomics [40,41], and additional annotated data from AIMI [38]. This model was selected because it was trained on a diverse and extensive collection of thoracic CT images containing NSCLC. nnUN-Net is a semantic segmentation method that automatically adapts to a given dataset by configuring a tailored U-Net-based segmentation pipeline.

### 2.2.2. Mediastinal lymph nodes

The mediastinal lymph nodes (MN) segmentation was conducted using a nnU-Net model, trained specifically for this task. Training data originated from the LNQ2023 MICCAI challenge, which comprises chest CT scans from 393 patients with lymphadenopathy with various cancer types, including breast cancer, NSCLC, renal cancer, and small cell lung cancer, among others[42]. In this public dataset, lymphadenopathy was specifically defined by the presence of clinically relevant lymph nodes larger than 1 cm in diameter.

### 2.2.3. Coronary arteries segmentation and coronary artery calcification scoring

Segmentation of the coronary arteries was achieved using the TotalSegmentator tool [43], specifically leveraging its dedicated coronary artery segmentation model suitable for non-contrast CT images. The coronary artery calcification (CAC) score was computed based on the established Agatston scoring [44–46]method: High-density regions (≥130 HU (Hounsfield unit)) were identified from the segmented coronary artery masks. On each axial slice, connected components were labeled and filtered to exclude calcium deposits smaller than 1 mm² in area. For each remaining calcium deposit (or plaque), the area was calculated and multiplied by a density-based weighting factor corresponding to its peak attenuation: 1 for 130-199 HU, 2 for 200-299 HU, 3 for 300-399 HU, and 4 for ≥400 HU. The CAC score was defined as the sum of weighted lesion scores across all slices.

Codes and segmentation scripts are available on our GitHub repository
https://github.com/shruti26mali/PixelsToPrognosis-NSCLC

## 2.3. Feature Extraction

### 2.3.1. Handcrafted radiomics features

Radiomic features were extracted with PyRadiomics [47] (version 3.0.1) based on the guidelines set by the Image Biomarker Standardization Initiative (IBSI) [48]. The parameters for extraction were the fixed bin width of 25, no intensity transformation, and no additional resampling of voxels. The extracted features belonged to mainly: (i) shape and volume-based features, which describe the geometric properties of the region of interest (ROI), (ii) first-order statistical features, which quantify the intensity distributions of the voxels in the ROI (e.g., man, variance, skewness, kurtosis, and entropy) and (iii) texture features, which capture spatial patterns and intensity heterogeneity within the ROI based on grey-level matrices. In total, 93 texture features and 14 shape/volume features were computed per applicable ROI. Specifically, texture features were derived from five matrix types: grey level co-occurrence matrix (GLCM), grey level run length matrix (GLRLM), grey level size zone matrix (GLSZM), neighbouring grey tone difference matrix (NGTDM), and grey level dependence matrix (GLDM). Shape and volume features were computed for ROIs with clear anatomical boundaries, such as the lung tumors and mediastinal lymph nodes, due to clinical relevance. No additional filters or image transformations were applied before feature extraction. This yielded a total of 107 features per ROI, including 14 shape and 93 texture features (i.e., 18 first-order and 75 higher-order). The extracted features per patient, selected according to the clinical relevance of each anatomical region, included: 93 texture features for the whole lung; 14 shape features and 93 texture features for the lung tumor; 14 shape features and 93 texture features for the mediastinal lymph nodes; and 93 texture features for the coronary arteries (feature extraction per patient).

### 2.3.2. Deep feature extraction using the Foundation model

Deep imaging features were extracted from the largest tumor region using a pretrained foundation model (FM) developed by Pai et al. [31]. For each patient, the largest tumor was identified, and isotropic image resampling (1 x 1 x 1 $mm^3$) voxel spacing was applied using B-spline interpolation, followed by CT intensity normalization consistent with the FM model suitable for lung CT scans. Intensity normalization involved clipping values between -1024 and 2048 HU and then normalizing these values to a [0,1] range. Subsequently, cubic patches centred on the tumor region were extracted in three different cube sizes (50, 96, and 128 voxels per side) to investigate size-dependent feature extraction performance. These patch sizes were selected to capture different spatial scales of tumor morphology and context. Model performance across cube sizes was compared to identify the most informative representation. The FM architecture incorporates a 3D ResNet-50 backbone for volumetric feature encoding and outputs a 4096-dimensional deep feature vector for each input cube.

## 2.4. Harmonization

To address centre-specific variability in imaging-derived features, we employed RKN for image-level harmonization and ComBat for feature-level harmonization.

### 2.4.1. Reconstruction Kernel Normalization (RKN):

Reconstruction kernel normalization (RKN) [25] addresses variability arising from different CT reconstruction kernels by standardizing the frequency content of CT images. The original CT image ($I_0$) is disbanded into a series of frequency components $F_i$ using Gaussian filters at multiple scales ($\sigma_i$ = 0, 1, 2, 4, 8, 16), producing filtered images $L_{\sigma_i}$. The frequency bands are computed as $F^{i+1} = L_{\sigma+1} - L_{\sigma_{i+1}}$ for $i = 0, 1, 2, 3, 4$ and $F^{i+1} = L_{\sigma_i}$ for $i = 5$ . The normalized image ($I_N$) is reconstructed by:

$$I_N = F^6 + \sum_{i=1}^{5} \lambda_i \, . F^5 \qquad (1)$$

Where $\lambda_i = \frac{r_i}{e_i}$, $r_i$ and $e_i$ represent the standard deviations of the frequency band $F_i$ in the reference image and original image $I_0$, respectively. This iterative process continues until all $\lambda_i$ fall within the range [0.95, 1.05]. In this study, we applied RKN as a preprocessing step to the entire thoracic CT image of each patient before radiomics feature extraction. Radiomic features were extracted from RKN-harmonized and original CT images for downstream analysis of lung and tumor models.

### 2.4.2. ComBat harmonization:

ComBat harmonization is an empirical Bayes statistical method originally developed to correct for batch effects in genomic data [49]. It models radiomic features according to:

$$y_{ij} = \alpha + \beta . X_{ij} + \gamma_i + \delta_i . \varepsilon_{ij} \qquad (2)$$

Where $y_{ij}$ is the radiomic feature for ROI $j$ on scanner $i, \alpha$ the average value for $y_{ij}$ $\beta$ captures the influence of biological covariates ($X_{ij}$ ), $\gamma_i$ and $\delta_i$ represents the additive and multiplicative scanner effect, respectively, and $\varepsilon_{ij}$ the error term. ComBat [50] adjusts for these scanner-induced batch effects while preserving biological variability. We applied ComBat harmonization, with batch effects from multiple sites, separately to texture features from images of original lung, RKN-harmonized lung, original tumor and RKN-harmonized tumor, original MN (mediastinal nodes), original CAC, and deep features from the foundation model for each cube size. The largest imaging centre with the most samples in the train set was chosen as the reference batch for ComBat harmonization.

## 2.5. Feature Selection

Once the radiomic features and deep features were extracted from all the ROIs, feature reduction was performed in a three-stage, cross-validated (stratified 5-fold) pipeline applied independently to each ROI (lung region, tumor, mediastinal nodes, coronary arteries) and the deep features extracted from the FM in order to prevent overfitting. The feature selection steps for all the models were as follows: (i) features that were constant or exhibited near-zero variance across the full training set were removed; (ii) highly correlated features were removed if the correlation ≥ 90% (ROI-only models) or ≥ 70% (clinical + ROI models or combination models). Eventually, features selected in more than 50% of the iterations were retained for subsequent survival analysis.

Since the FM-derived deep features were high-dimensional (4096 features), Principal Component Analysis (PCA) was employed to reduce dimensionality before model fitting. The number of PCA components was treated as a hyperparameter and optimized jointly with Cox model hyperparameters during survival model training.

## 2.6. Prognostic Model Construction

In survival analysis, the outcome of interest is time-to-event; here, it is overall survival (time from baseline to death or last follow-up). Conventional regression cannot model the combination of (i) right censoring (patient alive at last follow-up) and (ii) varying follow-up times; specialized survival models are required. We employed the Cox proportional-hazards (CoxPH) model, a semi-parametric approach that relates the hazard (instantaneous risk of death) to a linear combination of covariates without assuming a specific baseline-hazards shape. For a patient, the hazard function at time $t$ is:

$$h(t) = h_0(t). \exp(\sum_{i=1}^{n} \beta_i \, x_i) \qquad (3)$$

Where $h_0(t)$ is the baseline hazard function when all risk factors are absent ($x_i$ = 0), $h(t)$ is the hazard for the patient at time $t$, $x_i$ is the covariate vector, and $\beta_i$ are the log-hazard coefficients.

Models were fitted with *CoxPHfitter* from the *lifelines* Python package version 0.30.0, which implements the partial-likelihood estimator and allows elasticnet regularization to curb overfitting. Two hyperparameters, the global penalty and L1/L2 mixing factor, were optimized with Optuna (100 trials) inside a stratified five-fold cross-validation loop. The final model parameters were selected based on the average C-index on the validation set from cross-validation. Cox models were trained independently for each region of interest (ROI), namely, tumor, lungs, mediastinal nodes, and coronary arteries. In addition, we trained clinical-radiomic combination models, where clinical variables were concatenated with the selected radiomic features before modelling.

Following model training, patient-specific risk scores were generated for each patient using the predict_partial_hazard() function from the CoxPHfitter object. This method estimates the relative risk of experiencing the event based on the fitted model coefficients. Patients were then classified into high-risk and low-risk groups based on the median predicted risk score. Survival outcomes were visualized by plotting Kaplan-Meier (KM) survival curves for each risk group. To quantify the hazard between groups, we fit a univariable Cox model using this binary risk group (high vs. low) as the sole predictor.

To improve model interpretability, we employed SHapley Additive exPlanations (SHAP)[33] for Cox models to estimate the contribution of each feature to a patient's predicted risk. SHAP values were computed for both ROI-specific and combined models, allowing identification of the most influential features contributing to the prognostic signature.

## 2.7. Evaluation Metrics

Model performance was evaluated using the following metrics:

- **Concordance index (C-index)**

  The C-index measures the model's ability to correctly rank pairs of patients by relative risk. A value of 0.5 indicates random performance, and 1.0 indicates perfect discrimination. It was computed on both training and test sets using the lifelines implementation:

$$Concordance\ index = \frac{correct\ pairs + \frac{1}{2}.tied\ pairs}{all\ pairs} \quad (4)$$

  where *correct_pairs* are pairs where the patient with shorter survival time had a higher predicted risk, tied_pairs have equal risk scores, and all_pairs are all comparable pairs (i.e., not censored earlier). This formulation accounts for ties and censoring and is consistent with Harrell's C-index.

- **Time-dependent area under the ROC curve (AUC)**

  To assess discrimination at a fixed time point, we computed the time-dependent AUC at 5 years using cumulative_dynamic_auc from the scikit-survival package. This metric evaluates how well the model separates patients who experience the event before time t from those who survive beyond it. We passed the model's risk scores (from predict_partial_hazard) to the AUC function and evaluated at t = 5 years. To estimate confidence intervals (CI) and statistical significance, we applied bootstrap resampling (1,000 iterations). The 95 % CI and p-value were derived from the empirical distribution of AUC values across bootstrap samples.

- **Kaplan-Meier survival curves**

  Kaplan-Meier survival curves were generated for each (high or low risk) group, and survival differences were assessed using the log-rank test. In addition, a univariable Cox model using the binary risk group as a predictor was fit to report the hazard ratio (HR) with its 95% CI and p-value.

- **Consensus-based classification**

  To assess prediction robustness across anatomical regions, we implemented a strict consensus classification strategy using best-performing models (high C-index) from each ROI. At time horizons (2 or 5 years), we computed the predicted survival probability $S(t)$ for each test patient using the model's predict_survival_function() method. This function returns the model-estimated probability that a patient survives beyond time $t$, assuming entry at baseline (i.e., without conditioning on prior survival). Binary classification labels were assigned by thresholding $S(t)$ using a model-specific cutoff $\tau$, determined by maximizing Youden's index on the training set. We defined the predicted label $\hat{y}_i(t)$ for each patient $i$ at time $t$ as:

$$\hat{y}_i(t) = \begin{cases} 1, & if\ S_i(t) < \tau \\ 0, & if\ S_i(t) \geq \tau \end{cases} \quad (5)$$

  Where $S_i(t)$ is the survival probability for the patient $i$ at time $t$, and $\tau$ is the classification threshold. A patient was included in the consensus subset only if all selected ROI-specific models agreed on $\hat{y}_i(t)$. Consensus performance was evaluated using accuracy, sensitivity, specificity, and time-dependent AUC (t-AUC), and we also report consensus coverage (i.e., the proportion of valid test patients retained under strict agreement).

## 2.8. Radiomics Quality Score 2.0 assessment

The methodological quality of the proposed prognostic pipeline was assessed with the guidance of the Radiomics Quality Score (RQS 2.0) framework [51]. All of the RQS requirements, including data preparation, model development, model validation, and trustworthiness, were evaluated using the official scoring requirements. The cumulative score was mapped to the corresponding Radiomics Readiness Level (RRL) until level 6 to quantify methodological capability. Detailed scoring criteria and evidence mapping are provided in Supplementary Table 5.

## 2.9. Statistical Analysis

Appropriate statistical tests were used to compare variables between the training and test sets for the clinical characteristics (Table 2) and imaging parameters (Table 1). Continuous variables were compared using the independent t-test or the Mann-Whitney U test, based on normality. Categorical variables were compared using the Chi-squared test. Model performance was evaluated using the concordance index (C-index) and 5-year time-dependent AUC (t-AUC). Confidence intervals for both metrics were computed via 1,000-sample bootstrap resampling. For the t-AUC, a two-sided bootstrap test was used to assess significance, with p-values calculated as the proportion of t-AUC < 0.5 and 95% confidence intervals derived using the percentile method. Survival differences between high- and low-risk groups were assessed using the log-rank test, and a univariable Cox model was used to compute hazard ratios with 95% confidence intervals. Statistical significance was set at $p<0.05$.

# 3. Results

## 3.1. Data

A total of 876 patients with confirmed NSCLC and baseline thoracic CT scans were included in the final analysis, with 604 patients in the training set and 272 in the test set. The average age was similar between groups (64.7 ± 10.0 in train vs. 64.6 ± 9.9 in test, $p = 0.87$), with a slightly higher proportion of females in the test set (34.6%) compared to the training set (28.8%). No significant differences were observed in ECOG status, TNM staging, metastasis distribution, or survival time. However, a higher proportion of ex-smokers was present in the test set (45.6% vs. 35.3%, $p = 0.02$), and ECOG 1 status was more frequent in test patients (24.3% vs. 14.6%, $p = 0.049$).

Regarding imaging characteristics, the dataset included scans from five centres and six scanner manufacturers, with Philips and GE being the most prevalent. Centre distributions were imbalanced ($p < 0.001$), with CHU Angers contributing 29.8% of the test set versus 13.6% of the training set. Most scans were acquired at 120 kVp, and no significant differences were found in pixel spacing (mean: 0.67 mm in both sets) or slice thickness (mean: 1.5 mm in train vs. 1.48 mm in test, $p = 0.78$).
Full imaging acquisition parameters and clinical characteristics are reported in Tables 1 and 2.

Table 2. Baseline clinical characteristics of patients in the training and test sets. Variables include demographics, smoking history, PD-L1 expression, TNM staging, metastasis status by organ site, ECOG performance, survival status, and tumor histotype. Data are presented as mean ± standard deviation (SD) for continuous variables and as n (%) for categorical variables.

| Characteristic | Train (n=604) | Test (n=272) | p-value | Statistical comparison test |
|---|---|---|---|---|
| **Age** | | | | |
| Mean (SD) | 64.70 (10.04) | 64.57 (9.93) | 0.8653 | Ttest_ind |
| **Gender** | | | | |
| Female | 174 (28.8%) | 94 (34.6%) | 0.1031 | Chi-square |
| Male | 430 (71.2%) | 178 (65.4%) | | |
| **Packs year** | | | | |
| Available cases | 357 (59.10%) | 175 (28.07%) | | |
| NaNs cases | 247 (40.89%) | 97 (16.05%) | | |
| Mean (SD) | 45.37 (26.48) | 44.45 (36.85) | 0.2654 | mannwhitney |
| **Smoking status** | | | | |
| Non-smoker | 83 (13.7%) | 37 (13.6%) | 0.0227 | Chi-square |
| Ex-smoker | 213 (35.3%) | 124 (45.6%) | | |
| Smoker | 268 (44.4%) | 100 (36.8%) | | |
| NaN cases | 40 (6.6%) | 11 (4.0%) | | |
| **PDL1 expression value** | | | | |
| Available cases | 271 (44.86%) | 170 (28.14%) | | |
| NaNs cases | 333 (55.13%) | 102 (37.5%) | | |
| Mean (SD) | 31.80 (34.83) | 24.01 (32.86) | 0.0807 | mannwhitney |
| **Clinical stage group** | | | | |
| I | 95 (15.7%) | 39 (14.3%) | 0.6839 | Chi-square |
| II | 40 (6.6%) | 20 (7.4%) | | |
| III | 107 (17.7%) | 53 (19.5%) | | |
| IV | 210 (34.8%) | 83 (30.5%) | | |
| NaN cases | 152 (25.2%) | 77 (28.3%) | | |
| **ECOG performance status** | | | | |
| Grade 0 | 136 (22.5%) | 59 (21.7%) | 0.0488 | Chi-square |
| Grade 1 | 88 (14.6%) | 66 (24.3%) | | |
| Grade 2 | 19 (3.1%) | 15 (5.5%) | | |
| Grade 3 | 14 (2.3%) | 3 (1.1%) | | |
| Grade 4 | 4 (0.7%) | 3 (1.1%) | | |
| NaN cases | 343 (56.8%) | 126 (46.3%) | | |
| **event** | | | | |
| 0 (censored) | 286 (47.4%) | 117 (43.0%) | 0.2635 | Chi-square |
| 1 (death) | 318 (52.6%) | 155 (57.0%) | | |
| **Survival time (months)** | | | | |
| Mean (SD) | 28.68 (24.70) | 26.89 (23.64) | 0.3806 | mannwhitney |
| **Clinical metastasis staging** | | | | |
| cM0 | 263 (43.5%) | 122 (44.9%) | 0.5032 | |
| cM1 | 241 (39.9%) | 99 (36.4%) | | |
| NaN cases | 100 (16.6%) | 51 (18.8%) | | |
| **Clinical regional nodes staging** | | | | |
| cN0 | 159 (26.3%) | 67 (24.6%) | 0.3630 | Chi-square |
| cN1 | 45 (7.5%) | 24 (8.8%) | | |
| cN2 | 131 (21.7%) | 57 (21.0%) | | |
| cN3 | 123 (20.4%) | 44 (16.2%) | | |
| cNX | 12 (2.0%) | 10 (3.7%) | | |
| NaN cases | 134 (22.2%) | 70 (25.7%) | | |
| **Clinical tumor staging** | | | | |
| cT1 a/b/c | 104 (17.2%) | 52 (19.1%) | 0.2623 | Chi-square |
| cT2 a/b | 100 (16.6%) | 42 (15.4%) | | |
| cT3 | 95 (15.7%) | 43 (15.8%) | | |
| cT4 | 156 (25.8%) | 54 (19.9%) | | |
| cTX | 12 (2.0%) | 10 (3.7%) | | |
| NaN cases | 137 (22.7%) | 71 (26.1%) | | |
| **Personal cancer history** | | | | |
| No history | 113 (18.7%) | 42 (15.4%) | 0.2815 | Chi-square |
| History | 491 (81.3%) | 230 (84.6%) | | |
| **Tumor histotype** | | | | |
| Adenocarcinoma | 422 (69.9%) | 186 (68.4%) | 0.1233 | Chi-square |
| Squamous cell carcinoma | 126 (20.9%) | 47 (17.3%) | | |
| Non-small cell carcinoma | 51 (8.4%) | 35 (12.9%) | | |
| Large cell carcinoma | 5 (0.8%) | 4 (1.5%) | | |
| **adrenal gland metastasis** | | | | |
| No | 552 (91.4%) | 254 (93.4%) | 0.3836 | Chi-square |
| Yes | 52 (8.6%) | 18 (6.6%) | | |
| **Bone metastasis** | | | | |
| No | 497 (82.3%) | 216 (79.4%) | 0.3590 | Chi-square |
| Yes | 107 (17.7%) | 56 (20.6%) | | |
| **Brain metastasis** | | | | |
| No | 518 (85.8%) | 236 (86.8%) | 0.7708 | Chi-square |
| Yes | 86 (14.2%) | 36 (13.2%) | | |
| **Liver metastasis** | | | | |
| No | 553 (91.6%) | 249 (91.5%) | 1.0000 | Chi-square |
| Yes | 51 (8.4%) | 23 (8.5%) | | |
| **Lung metastasis** | | | | |
| No | 521 (86.3%) | 228 (83.8%) | 0.3990 | Chi-square |
| Yes | 83 (13.7%) | 44 (16.2%) | | |
| **Lymph nodes metastasis** | | | | |
| No | 509 (84.3%) | 225 (82.7%) | 0.6332 | Chi-square |
| Yes | 95 (15.7%) | 47 (17.3%) | | |
| **Muscle metastasis** | | | | |
| No | 594 (98.3%) | 269 (98.9%%) | 0.7459 | Chi-square |
| Yes | 10 (1.7%) | 3 (1.1%) | | |
| **Pleura metastasis** | | | | |
| No | 579 (95.9%) | 254 (93.4%) | 0.1609 | Chi-square |
| Yes | 25 (4.1%) | 18 (6.6%) | | |
| **Other metastasis** | | | | |
| No | 578 (95.7%) | 257 (94.5%) | 0.5407 | Chi-square |
| Yes | 26 (4.3%) | 15 (5.5%) | | |
| | | | | |

## 3.2. Segmentation

As shown in Figure 3, ROI segmentations captured a wide spectrum of disease presentation, including tumor burden, nodal involvement, and coronary calcification.

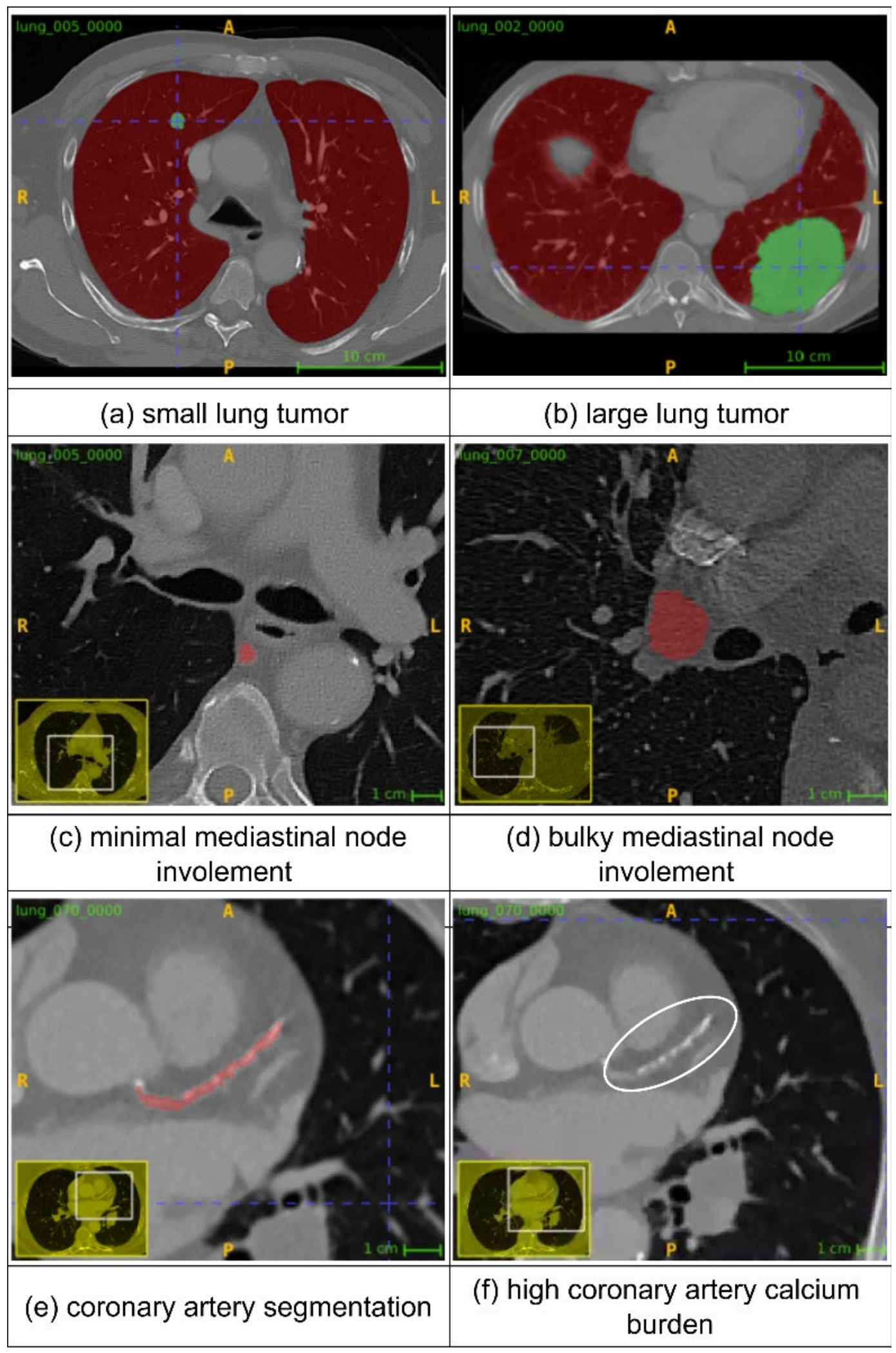

Figure 3. Representative CT slices illustrating segmentation and variability across anatomical regions of interest (ROIs). (a-b) Axial views of the segmented tumor (red) and lung regions. (a) shows a patient with a small tumor; (b) shows a large tumor occupying most of the left lung. (c-d) Axial views of mediastinal node (MN) segmentation (red). (c) Illustrates minimal nodal involvement; (d) shows bulky nodal disease near the main bronchi. (e-f) Axial views of coronary artery (CA) segmentation (e, red) and corresponding calcium burden (f, circled in white) in a patient with a high CAC score.

The mean tumor volumes (largest tumor) and mean MN volumes were analysed against clinical staging categories (refer Figure 4). As expected the tumor volume increased gradually with higher T-, N-, and M-staging which showed tumor burden in advancing lung cancer stages. MN volumes also increased with higher N-staging showing that the enlargement of regional nodes implied nodal involvement and disease spread. These findings reflect the biological consistency of the tumor and nodal annotations used for feature extraction.

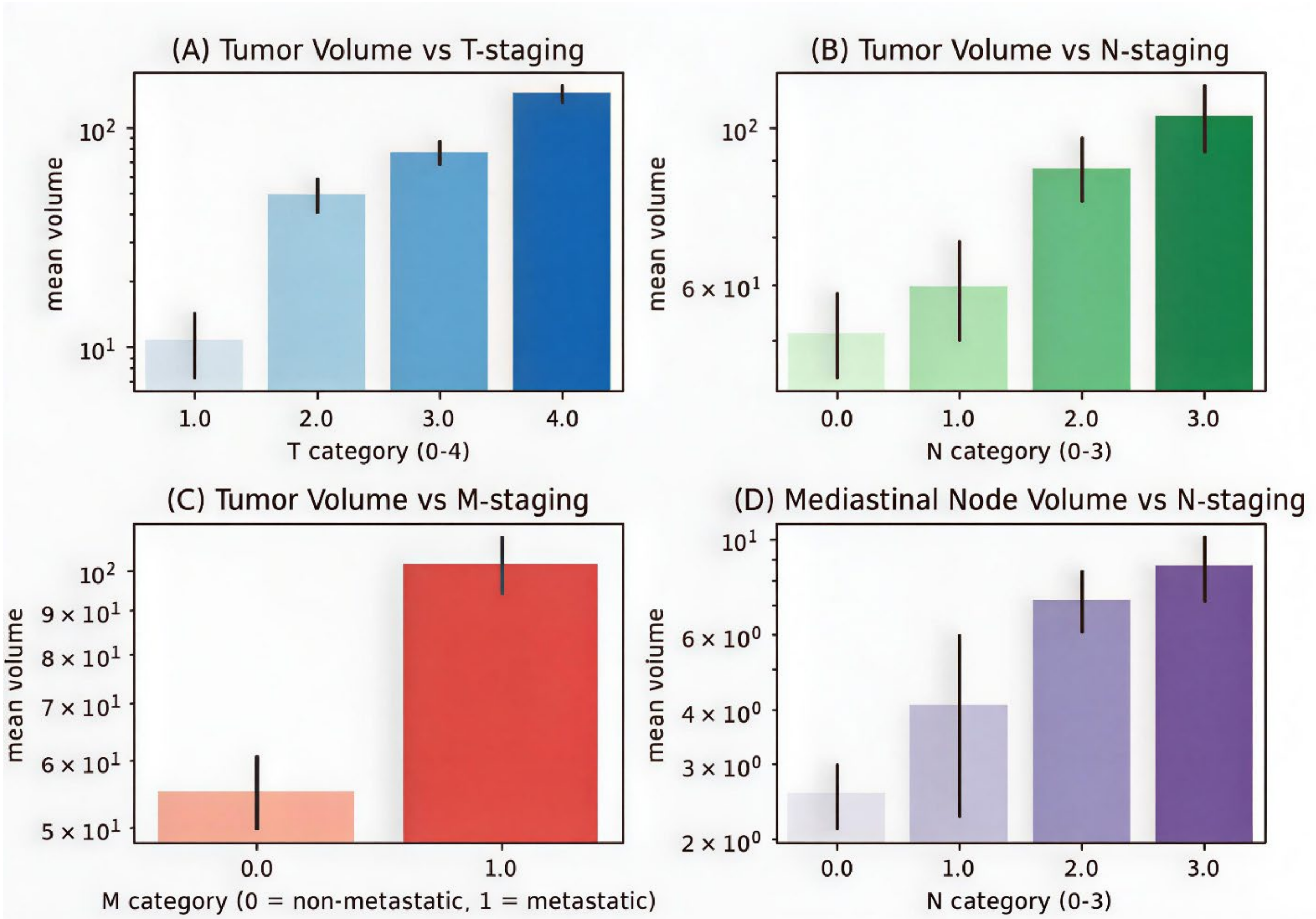

Figure 4. Mean tumor and mediastinal node volumes across clinical staging categories.
(A) Tumor volume increased with higher T-staging, reflecting greater local tumor burden.
(B) Tumor volume increased with advancing N-staging, suggesting association between primary tumor size and regional spread.
(C) Metastatic patients showed higher mean tumor volumes compared to non-metastatic cases.
(D) MN volumes increased with N-staging, indicating that regional node size corresponded to disease progression.

To evaluate the prognostic models of different anatomical regions (ROIs), results are organized per region of interest (ROI), including the whole lung region, tumor, mediastinal nodes, coronary arteries (CA), and coronary artery calcium (CAC) score, and FMCIB features (refer Table 3). The table also consists of models where ROI features were harmonized appropriately (see before and after harmonization effects), and compared against models per ROI.

Table 3: Survival analysis of models built from clinical variables, handcrafted radiomic features across multiple anatomical regions (whole lung region, tumor, mediastinal lymph nodes, coronary arteries (CA) , and coronary artery calcium (CAC) score), and foundation model (FM) deep features; before and after harmonization. Reported metrics include concordance index (C-index), hazard ratio (HR) with 95% confidence interval (CI), log-rank test p-values, and 5-year time-dependent area under the curve (t-AUC). ComBat, reconstruction kernel normalization (RKN), and their combination (RKN+ComBat) were applied where relevant. Best-performing models per category are highlighted in bold.

| **Model** | **Harmonization** | **Test (C-index)** | **Test (Hazard ratio [CI 95%]))** | **p-value (KM) TEST** | **Test (AUC at T = 5 years)** |
|---|---|---|---|---|---|
| **CLINICAL VARIABLES** | | | | | |
| Clinical model (diagnostic variables) (nTr=604, nTs=272) | - | **0.73 [0.69-0.77]** | **1.87 [1.35-2.58]** | **0.0001** | **0.88 [0.80-0.94] P=0.0000** |
| Sub-group (M0) (no metastatic vars) (nTr=363, nTs=173) | - | 0.72 [0.66-0.77] | 0.94 [0.57-1.54] | 0.8073 | 0.88 [0.80-0.94] P=0.0000 |
| Sub-group (M1 / M1a / M1b / M1c) (nTr=241, nTs=99) | - | 0.66 [0.59-0.72] | 0.93 [0.52-1.63] | 0.7883 | 0.69 [0.37-0.93] P=0.1831 |
| Metastasis indicator only (yes/no) (nTr=504, nTs=221) | - | 0.67 [0.63-0.72] | 2.29 [1.61-3.26] | 0.0000 | 0.73 [0.65-0.81] P=0.0000 |
| TNM staging (diagnostic variables) (nTr=604, nTs=272) | - | 0.67 [0.63-0.71] | 2.70 [1.94-3.75] | 0.0000 | 0.85 [0.77-0.92] P=0.0000 |
| **WHOLE LUNG REGION** | | | | | |
| Lung (texture) | - | 0.63 [0.59-0.68] | 1.87 [1.36-2.58] | 0.0001 | 0.62 [0.52-0.72] P=0.0020 |
| Lung (texture) | ComBat | **0.65 [0.60-0.69]** | **1.95 [1.4-2.7]** | **0.0000** | **0.65 [0.54-0.75] P=0.0020** |
| Lung (texture) | RKN | 0.62 [0.58-0.67] | 1.75 [1.27-2.40] | 0.0005 | 0.65 [0.54-0.75] P=0.0020 |
| Lung (texture) | RKN + ComBat | 0.63 [0.58-0.67] | 1.86 [1.35-2.57] | 0.0001 | 0.65 [0.55-0.75 P=0.0020 |
| Clinical + Lung (texture) | - | 0.75 [0.70-0.78] | 4.38 [3.09-6.2] | 0.0000 | 0.87 [0.8-0.93] P=0.0000 |
| Clinical + Lung (texture) | ComBat | **0.75 [0.70-0.78]** | **4.99 [3.5-7.13]** | **0.0000** | **0.87 [0.8-0.93] P=0.0000** |
| Clinical + Lung (texture) | RKN | 0.75 [0.72-0.79] | 4.54 [3.2-6.46] | 0.0000 | 0.86 [0.8-0.92] P=0.0000 |
| Clinical + Lung (texture) | RKN + ComBat | 0.74 [0.7-0.77] | 4.30 [3.04-6.09] | 0.0000 | 0.89 [0.83-0.95] P=0.0000 |
| **TUMOR** | | | | | |
| Tumor (volume) | - | 0.63 [0.58-0.67] | 1.70 [1.24-2.34] | 0.0009 | 0.67 [0.57-0.78] P=0.0020 |
| Tumor (texture) | - | 0.67 [0.67-0.72] | 2.23 [1.61-3.08] | 0.0000 | 0.73 [0.63-0.83] P=0.0000 |
| Tumor (texture) | ComBat | **0.69 [0.65-0.73]** | **3.68 [2.6-5.21]** | **0.0000** | **0.75 [0.65-0.84] P=0.0000** |
| Tumor (texture) | RKN | 0.67 [0.63-0.71] | 2.55 [1.84-3.53] | 0.0000 | 0.72 [0.62-0.81] P=0.0000 |
| Tumor (texture) | RKN + ComBat | 0.69 [0.64-0.73] | 3.48 [2.47-4.91] | 0.0000 | 0.76 [0.66-0.85] P=0.0020 |
| Clinical + Tumor (volume) | - | 0.75 [0.71-0.78] | 4.11 [2.91-5.79] | 0.0000 | 0.87 [0.81-0.93] P=0.0000 |
| Clinical + Tumor (texture) | - | 0.75 [0.72-0.79] | 4.80 [3.36-6.84] | 0.0000 | 0.87 [0.8-0.93] P=0.0000 |
| Clinical + Tumor (texture) | ComBat | **0.76 [0.72-0.79]** | **4.33 [3.05-6.14]** | **0.0000** | **0.88 [0.81-0.94] P=0.0000** |
| Clinical + Tumor (texture) | RKN | 0.75 [0.71-0.78] | 4.05 [2.88-5.69] | 0.0000 | 0.86 [0.8-0.92] P=0.0000 |
| Clinical + Tumor (texture) | RKN + ComBat | 0.76 [0.72-0.79] | 4.32 [3.05-6.13] | 0.0000 | 0.88 [0.82-0.93] P=0.0000 |
| **MEDIASTINAL NODES** | | | | | |
| MN (volume) | - | 0.57 [0.52-0.61] | 1.11 [0.78-1.59] | 0.5558 | 0.55 [0.43-0.68] P=0.4200 |
| MN (texture) | - | 0.56 [0.51-0.60] | 1.10 [0.77-1.56] | 0.620 | 0.57 [0.46-0.69] P=0.2460 |
| MN (texture) | ComBat | **0.62 [0.58-0.67]** | **1.29 [0.93-1.79]** | **0.1210** | **0.66 [0.55-0.77] P=0.0060** |
| Clinical + MN (volume) | - | 0.75 [0.72-0.79] | 4.02 [2.84-5.68] | 0.0000 | 0.9 [0.83-0.95] P=0.0000 |
| Clinical + MN (texture) | - | 0.74 [0.7-0.78] | 3.49 [2.48-4.90] | 0.0000 | 0.86 [0.79-0.93] P=0.0000 |
| Clinical + MN (texture) | ComBat | **0.76 [0.72-0.8]** | **4.24 [3.0-6.01]** | **0.0000** | **0.86 [0.78-0.93] P=0.0000** |
| **CORONARY ARTERIES & CAC SCORE** | | | | | |
| CA (texture) | - | 0.58 [0.53-0.62] | 1.20 [0.88-1.65] | 0.2569 | 0.59 [0.46-0.71] P=0.1440 |
| CA (texture) | ComBat | 0.58 [0.53-0.63] | 1.54 [1.12-2.12] | 0.0074 | 0.60 [0.47-0.73] P=0.1140 |
| CAC score | - | 0.6 [0.55-0.64] | 1.48 [1.08-2.03] | 0.0155 | 0.59 [0.45-0.72] P=0.1800 |
| Clinical + CA (texture) | - | 0.74 [0.7-0.77] | 4.39 [3.08-6.24] | 0.0000 | 0.86 [0.77-0.93] P=0.0000 |
| Clinical + CA (texture) | ComBat | **0.75 [0.71-0.78]** | **3.83 [2.70-5.40]** | **0.0000** | **0.9 [0.84-0.95] P=0.0000** |
| Clinical + CAC score | - | **0.75 [0.71-0.79]** | **4.23 [2.97 -6.01]** | **0.0000** | **0.88 [0.82-0.93] P=0.0000** |
| **FMCIB DEEP FEATURES** | | | | | |
| FMCIB (cube size = 128) | - | 0.65 [0.61-0.69] | 1.99 [1.44-2.74] | 0.0000 | 0.65 [0.55-0.74] P=0.0040 |
| FMCIB (cube size = 96) | - | 0.51 [0.46-0.56] | 2.21 [1.6-3.05] | 0.0000 | 0.65 [0.54-0.74] P=0.0060 |
| FMCIB (cube size = 50) | - | 0.66 [0.61-0.70] | 2.21 [1.6-3.05] | 0.0000 | 0.66 [0.56-0.76] P=0.0020 |
| FMCIB (cube size = 128) | ComBat | **0.67 [0.63-0.72]** | **2.73 [1.95-3.82]** | **0.0000** | **0.72 [0.63-0.81] P=0.0000** |
| FMCIB (cube size = 96) | ComBat | 0.43 [0.38-0.48] | 1.04 [0.76-1.43] | 0.8069 | 0.45 [0.35-0.56] P=1.7260 |
| FMCIB (cube size = 50) | ComBat | **0.67 [0.63-0.72]** | **2.55 [1.84-3.55]** | **0.0000** | **0.74 [0.65-0.83] P=0.0000** |
| Clinical + FMCIB (cube size = 128) | - | 0.75 [0.72-0.79] | 5.31 [3.71-7.59] | 0.0000 | 0.88 [0.81-0.94] P=0.0000 |
| Clinical + FMCIB (cube size = 96) | - | 0.75 [0.72-0.79] | 1.99 [1.43-2.74] | 0.0000 | 0.76 [0.67-0.85] P=0.0000 |
| Clinical + FMCIB (cube size = 50) | - | 0.76 [0.72-0.8] | 4.89 [3.42-6.97] | 0.0000 | 0.9 [0.84-0.95] P=0.0000 |
| Clinical + FMCIB (cube size = 128) | ComBat | **0.75 [0.71-0.79]** | **5.01 [3.50-7.15]** | **0.0000** | **0.89 [0.81-0.95] P=0.0000** |
| Clinical + FMCIB (cube size = 96) | ComBat | 0.57 [0.51-0.62] | 1.87 [1.35-2.58] | 0.0001 | 0.64 [0.55-0.73] P=0.0060 |
| Clinical + FMCIB (cube size = 50) | ComBat | **0.76 [0.73-0.8]** | **4.75 [3.33-6.79]** | **0.0000** | **0.89 [0.82-0.94] P=0.0000** |

## 3.3. Clinical models

The full clinical integrated model, incorporating diagnostic and demographic variables, achieved a C-index of 0.73 (95% CI: 0.69-0.77) and a 5-year t-AUC of 0.88 (95% CI: 0.80-0.94) on the test set (refer Table 3 CLINICAL VARIABLES subsection). The model stratified patients into high- and low-risk survival groups with a hazard ratio (HR) of 1.87 (95% CI: 1.35-2.58, p = 0.0001). The corresponding Kaplan-Meier survival curves for this stratification are shown in Figure 5.

Subgroup analyses by metastasis status showed diverging performance. In the M0 subgroup (patients without metastatic variables), the model maintained good discrimination (C-index = 0.72; t-AUC = 0.88), but the HR was not statistically significant (HR = 0.94, p = 0.81), indicating limited survival separation within this group. The M1 subgroup (with distant metastases) similarly showed modest discrimination (C-index = 0.66) and an HR of 0.93 (p = 0.79), with poor KM separation and wide confidence intervals.

Additional simplified models using only the M-staging or TNM stage categories still achieved meaningful prognostic performance. The M-staging model reached a C-index of 0.67, HR of 2.29 (95% CI: 1.61-3.26), and t-AUC of 0.73, while the TNM staging model achieved similar results (C-index = 0.67; HR = 2.70; t-AUC = 0.85).

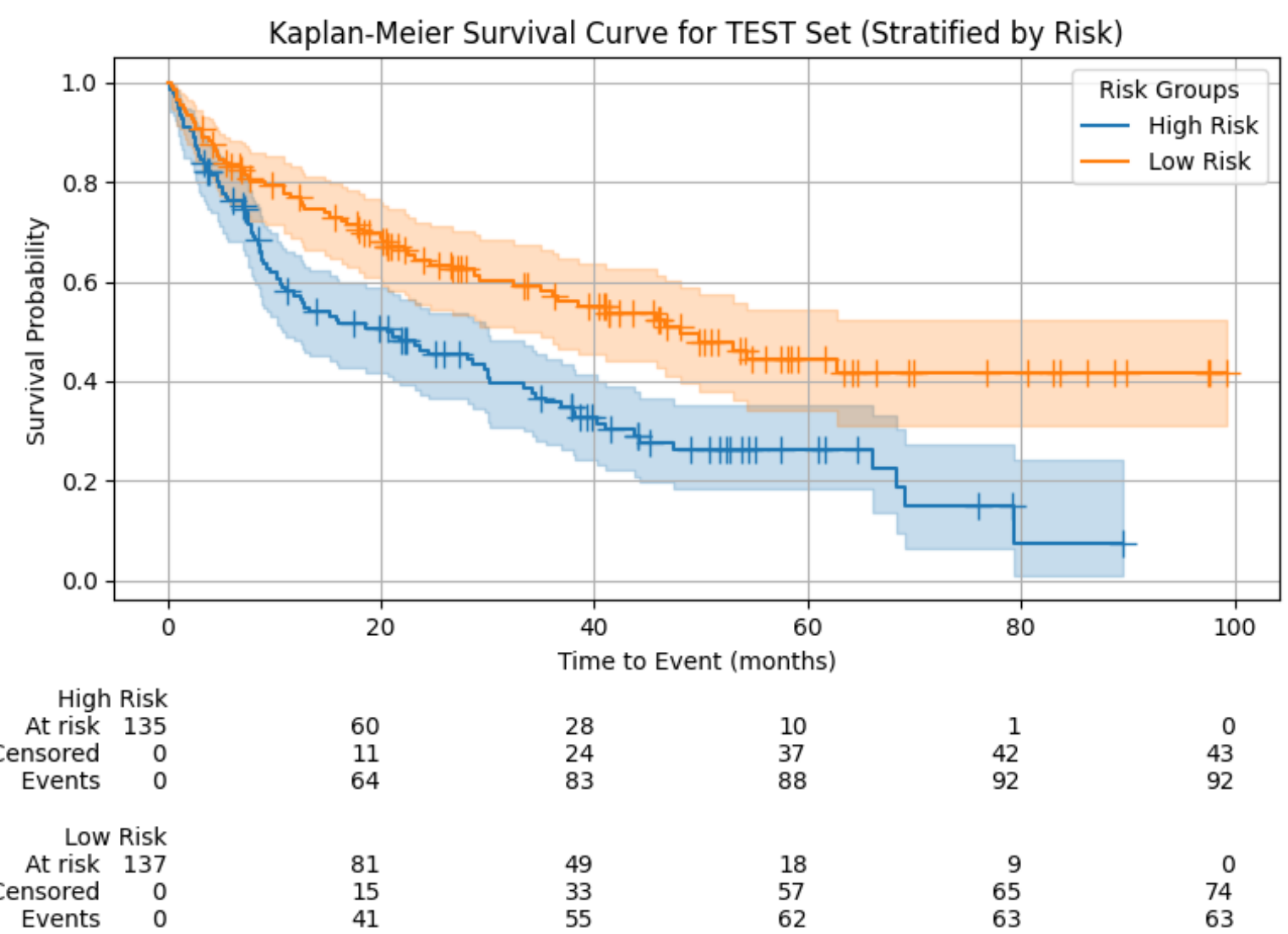

Figure 5. Kaplan-Meier curves computed on the test set for the full clinical model, incorporating all diagnostic and demographic variables.

## 3.4. Whole lung region

As shown in Table 3 WHOLE LUNG REGION subsection, The whole lung texture features provided moderate prognostic discrimination. The unharmonized model achieved a C-index of 0.63 and HR = 1.87, which improved following ComBat harmonization (C-index = 0.65, HR = 1.95, t-AUC = 0.65). RKN-only model showed no improvement (C-index = 0.62, HR = 1.75) whereas RKN + ComBat achieved comparable results (C-index = 0.63 , HR = 1.86). Combining clinical variables with lung lung texture features enhanced prognostic performance with a C-index of 0.75 and a 5-year t-AUC of 0.87-0.89 where ComBat-harmonized models consistently outperformed unharmonized ones. These findings suggest that background parenchymal changes contribute independently to survival risk stratification.

## 3.5. Tumor region

As shown in Table 3 TUMOR subsection, the model trained on tumor texture features achieved the strongest performance among radiomics-only (no harmonization models, with a C-index of 0.67, a 5-year t-AUC of 0.73 (95% CI: 0.63-0.83, p = 0.0000), and a hazard ratio (HR) of 2.23 (95% CI: 1.61-3.08). The corresponding Kaplan-Meier (KM) curve (Figure 6) illustrates clear stratification between the predicted high- and low-risk groups on the test set. The tumor volume (shape features) model also demonstrated modest predictive ability (C-index = 0.63, t-AUC = 0.67), suggesting that tumor burden contributes independently to survival risk stratification.

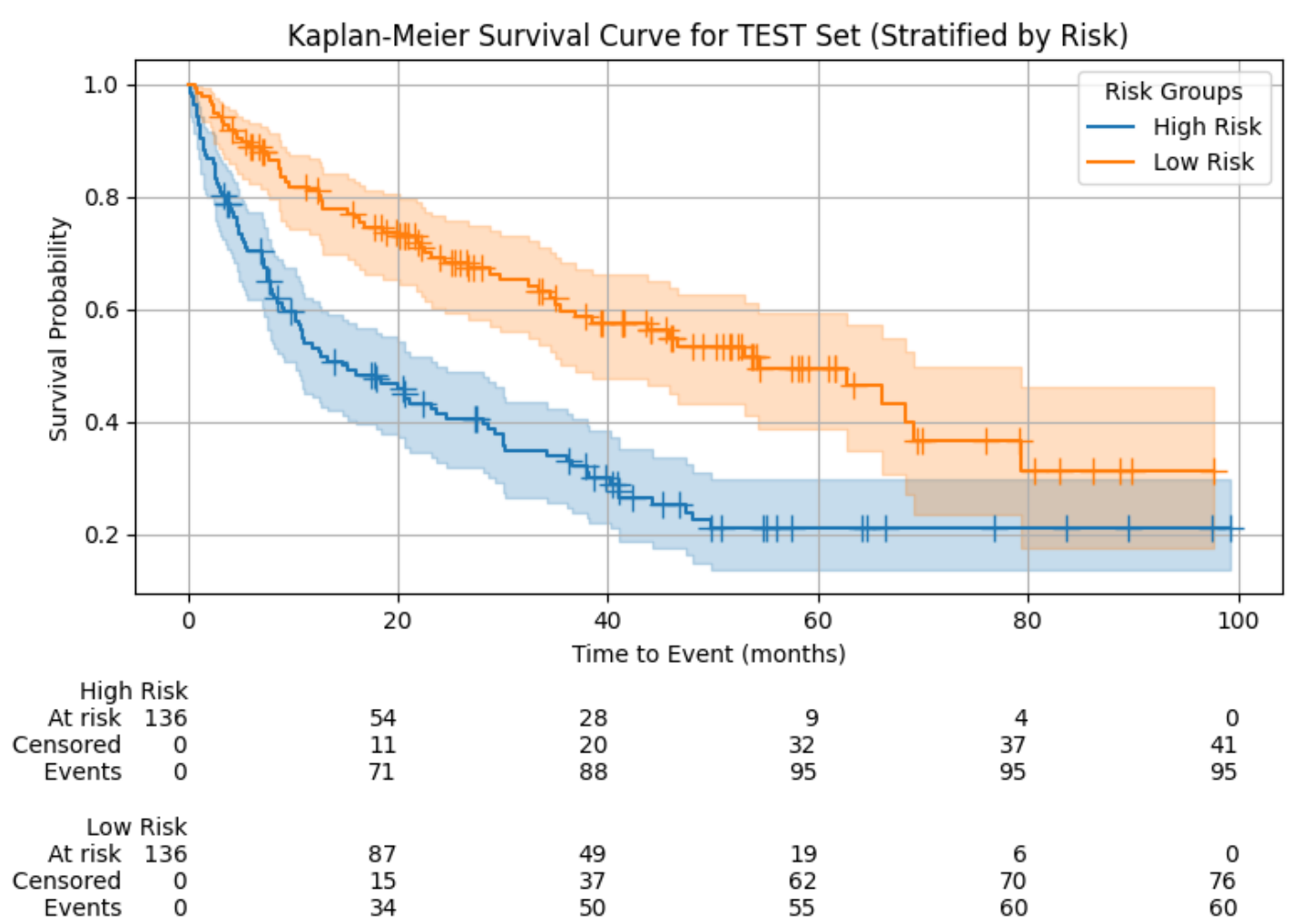

Figure 6. Kaplan-Meier curves computed on the test set for the tumor texture model (no harmonization) with HR of 2.23 (95% CI: 1.61-3.08) and log-rank p=0.0000 reflecting significant survival differences between high- and low-risk groups

We evaluated survival models trained using radiomic features harmonized via ComBat, RKN, or their combination. As shown in Table 3, the best-performing tumor only model was the tumor texture model harmonized with ComBat, which achieved a C-index of 0.69 (95% CI: 0.65-0.73), HR of 3.68 (95% CI: 2.6-5.21), and a 5-year t-AUC of 0.75 (95% CI: 0.65-0.84, p = 0.0000), outperforming RKN and RKN + ComBat variants especially with respect to HR. The RKN + ComBat combination on tumor texture also performed well, with a C-index of 0.69, HR = 3.48 (95% CI: 2.47-4.91), and t-AUC = 0.76 (95% CI: 0.66-0.85, p = 0.0020). The RKN-only version was slightly lower in performance (C-index = 0.67, t-AUC = 0.72). The calibration analysis (Appendix Figure 1) showed that the tumor texture model with ComBat harmonized features was well aligned with the observed 5-year survival probabilities. The curve closely followed the ideal reference line, showing good overall calibration and reliable risk estimation across the test cohort. Combining radiomic features with clinical variables consistently improved survival model performance across all ROIs. As shown in Table 3, the clinical + tumor texture model (no harmonization applied) achieved a high C-index (0.75) and strong survival separation (HR = 4.80, 95% CI: 3.36-6.84; p = 0.000), with a t-AUC of 0.87 (95% CI: 0.8-0.93).

We further evaluated the effect of harmonization on clinical + tumor models. Integrating clinical variables with tumor texture features further improved discrimination highlighting the complementary prognostic value of radiomic descriptors combined with clinical variables. Among clinical + radiomics models, the ComBat-harmonized tumor texture model performed strongly (C-index = 0.76, t-AUC = 0.88, HR = 4.33, 95% CI: 3.05-6.14), while RKN and RKN+ComBat variants performed comparably (C-index = 0.75-0.76, t-AUC = 0.86-0.88). The Kaplan-Meier plot in Figure 7 further demonstrates clear stratification between predicted high- and low-risk groups for clinical + tumor texture (ComBat harmonization). The tumor volume model along with clinical variables achieved similar results (C-index = 0.75, HR = 4.11, t-AUC = 0.87), indicating that both tumor burden and tumor texture heterogeneity independently contribute to the patient risk stratification.

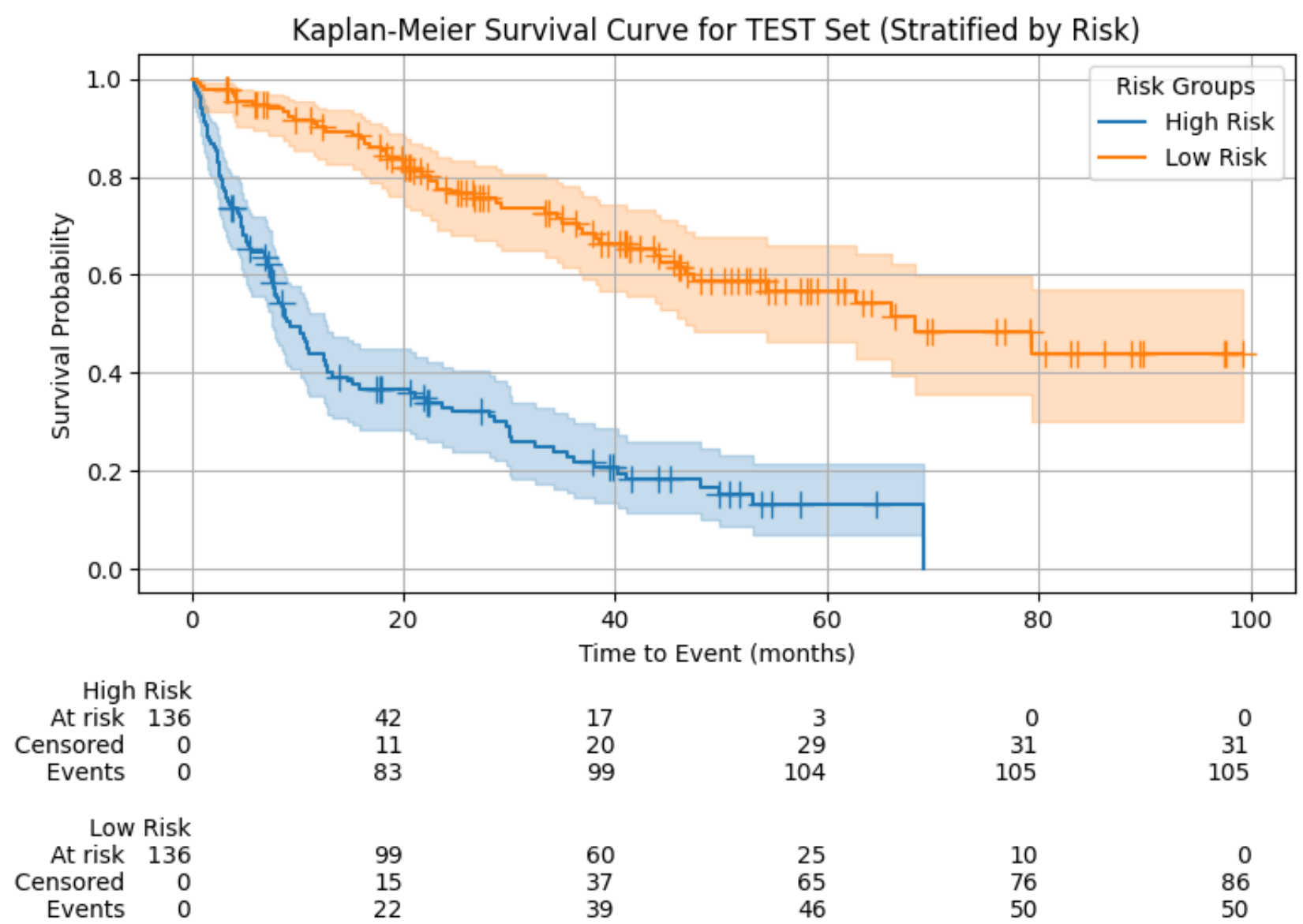

Figure 7: Kaplan-Meier curves computed on the test set for the clinical + tumor texture models with ComBat harmonization achieving HR of 4.80, 95% CI: 3.36-6.84 and log-rank p=0.0000 reflecting significant survival differences between high- and low-risk groups

## 3.6. Mediastinal nodes

As shown in Table 3 MEDIASTINAL NODES subsection, models trained on mediastinal node (MN) texture features showed lower discrimination (C-index=0.56) and lacked statistically significant survival separation (log-rank $p > 0.05$). A similar trend was observed for MN volume mode with C-index = 0.57, HR of 1.11 and poorly stratified risk groups with p = 0.5558. The ComBat-harmonized MN texture model showed modest discrimination (C-index = 0.62), with an HR of 1.29 (95% CI: 0.93-1.79) and t-AUC of 0.66 (95% CI: 0.55-0.77, p = 0.0060), although the KM p-value was non-significant (p = 0.1210). In contrast, combining MN features with clinical variables improved prognostic performance. The clinical + MN texture model with ComBat achieved C-index = 0.76 (0.72-0.80), HR = 4.24 (3.06-6.01), and t-AUC = 0.86 (0.78-0.93) (p = 0.0000), matching the clinical + tumor texture performance. The clinical + MN volume model also achieved a comparable C-index of 0.75 and a high t-AUC of 0.9. These findings suggest that radiomic descriptors from mediastinal lymph nodes, particularly after harmonization, capture complementary regional disease characteristics relevant to patient survival.

## 3.7. Coronary arteries and CAC score

As shown in Table 3 CORONARY ARTERIES & CA SCORE subsection, coronary artery (CA) and coronary artery calcium (CAC) features exhibited weaker individual prognostic power compared with tumor features. The CA texture model (without harmonization) showed low discrimination (C-index = 0.58, HR = 1.20 (0.88-1.65), t-AUC = 0.59, lacked statistically significant survival separation (log-rank p > 0.05)), but ComBat harmonization did not improve the results (C-index = 0.58, HR = 1.54 (1.12-2.12), t-AUC = 0.60, KM test p = 0.0074). The coronary artery calcium (CAC) score yielded a C-index of 0.6 and HR of 1.48 (95% CI: 1.08-2.03), with significant stratification in KM analysis (p = 0.02), suggesting that while CAC may reflect cardiovascular comorbidity, it can independently stratify cancer-specific survival. However, when combined with clinical variables, both CA texture and CAC score improved survival discrimination (C-index = 0.75, t-AUC = 0.88-0.90), underscoring their additive prognostic value through cardiovascular comorbidity information.

## 3.8. Foundation model deep features

As shown in Table 3 FMCIB DEEP FEATURES subsection, FM deep features extracted from 3D tumor patches also demonstrated prognostic value. The cube size = 50 achieved a C-index of 0.66, a t-AUC of 0.66 (95% CI: 0.56-0.76, p = 0.0020), and an HR of 2.21 (95% CI: 1.6-3.05, log rank p = 0.0000). The KM curve for this model (Figure 8) also shows clear separation between risk groups. Other FM cube sizes (96 and 128) yielded consistent performance (C-index range: 0.51-0.65), confirming the stability of FM-based feature representations across patch scales. While the FM-128 model aligns with the 95th percentile tumor size (Appendix Table 1), the FM-50 model appears to better capture prognostically relevant intra-tumoral heterogeneity.

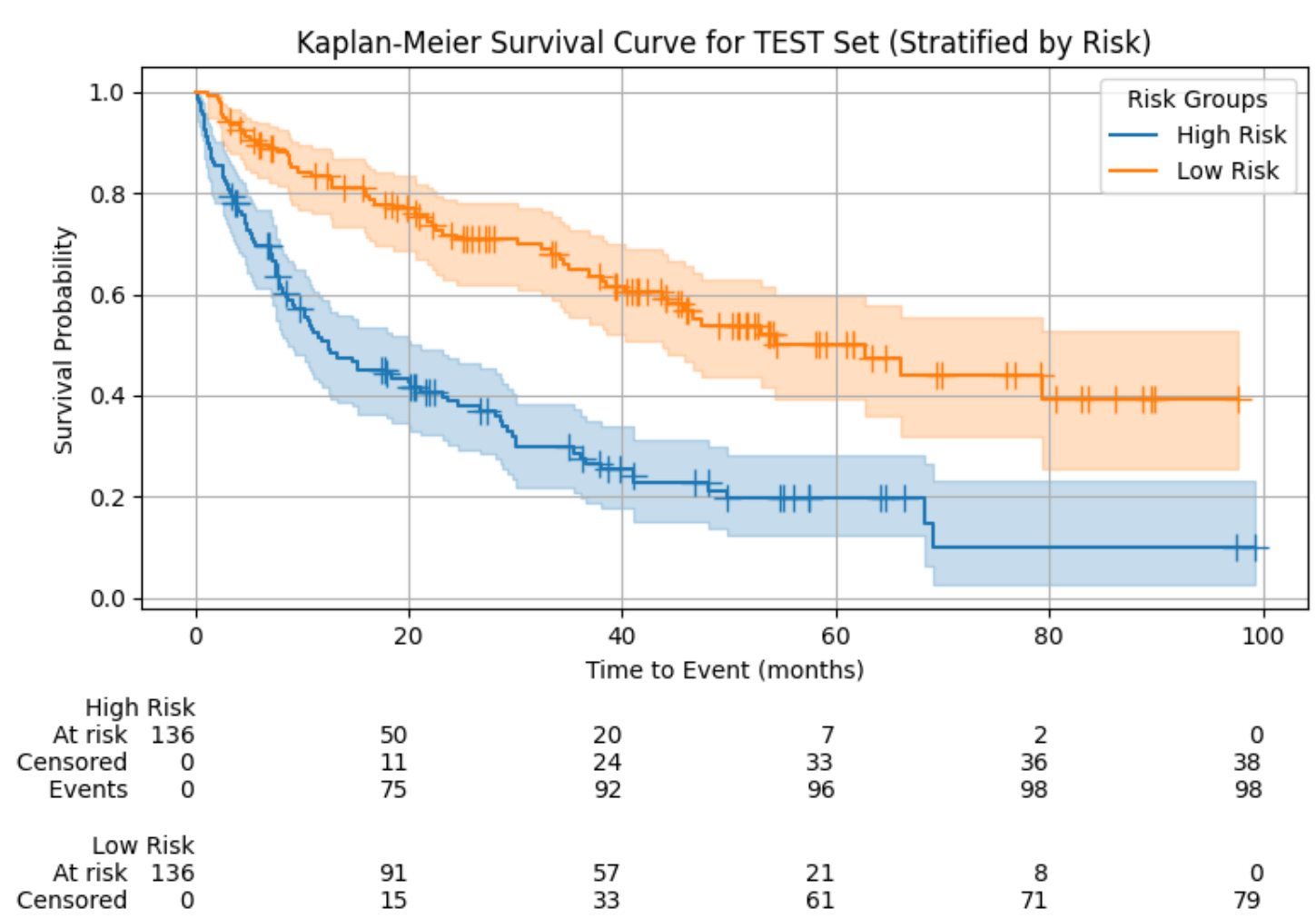

Figure 8. Kaplan-Meier curves computed on the test set for the FM deep feature (cube size = 50) survival model with HR of 2.21 (95% CI: 1.6-3.05) and log-rank p=0.0000 reflecting significant survival

Among ComBat-harmonized models, cube size = 50 and 128 achieved comparable performance with a C-index of 0.67, HR range: 2.55-2.73, and t-AUC: 0.72-0.74, while the cube size = 96 model showed poor discrimination after harmonization. Incorporating clinical variables along with harmonization further enhanced discrimination, the cube size = 128 model (no harmonization) also achieved the highest hazard ratio (HR = 5.31, 95% CI: 3.71-7.59) and strong discriminative performance (C-index = 0.75, t-AUC = 0.88). The cube size = 50 model slightly outperformed in C-index (0.76) and t-AUC (0.9, 95% CI: 0.84-0.95), indicating consistent prognostic power of FM-derived features across spatial scales. The corresponding KM curves (Figure 9) illustrate effective separation for the FM-128 model as well. And the clinical + FMCIB (cube = 50, ComBat) model also achieved a high overall prognostic performance (C-index = 0.76 (0.73-0.80), HR = 4.75 (3.33-6.79), t-AUC = 0.89 (0.82–0.94)).

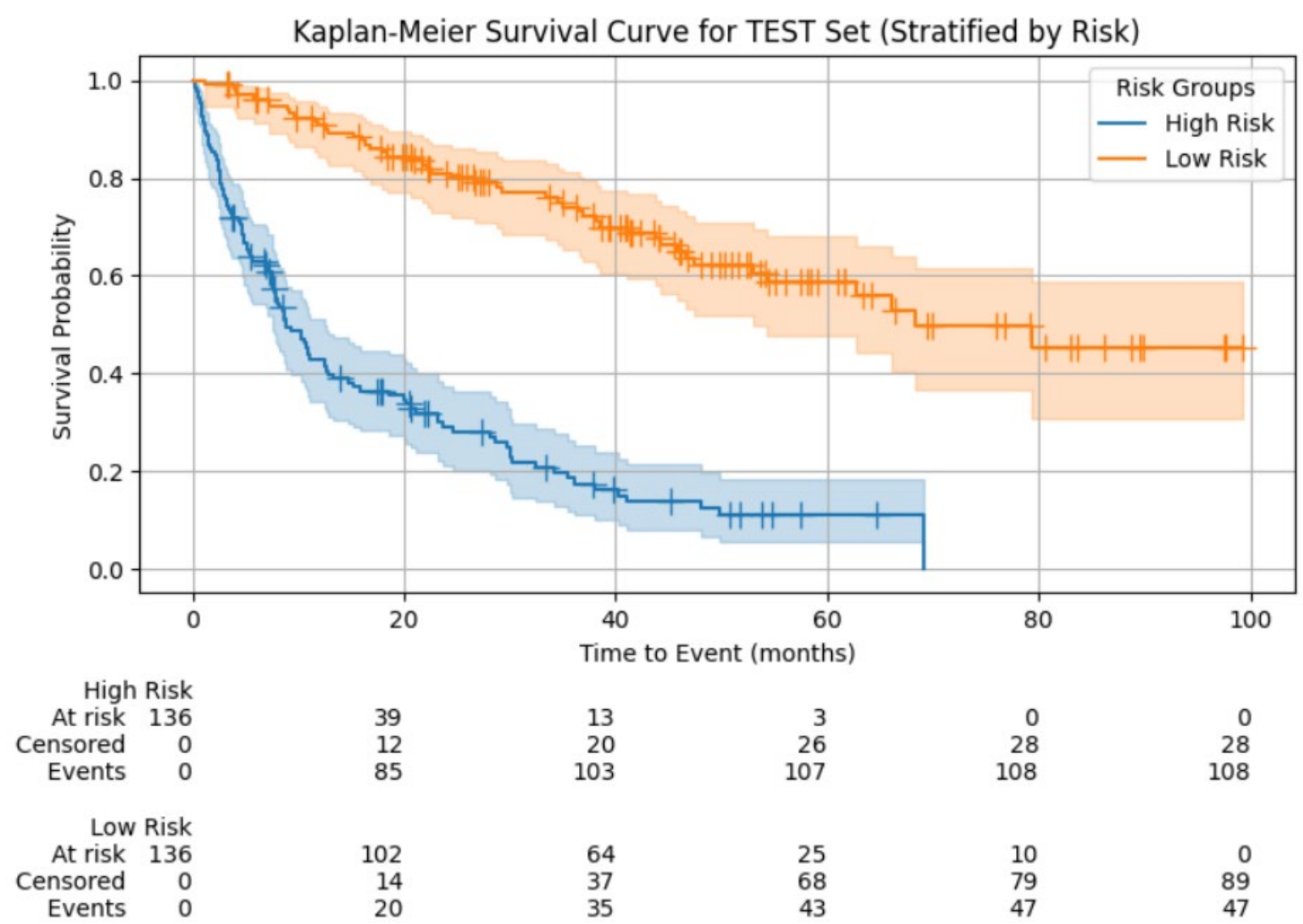

Figure 9. Kaplan-Meier curves computed on the test set for the clinical+FM deep feature (cube size = 128, no harmonization) survival model with HR of 5.31 (95% CI: 3.71-7.59) and log-rank p=0.0000 reflecting significant survival differences between high- and low-risk groups

## 3.9. Explainability - SHAP analysis

Figure 10 shows the SHAP summary plot for the clinical-only model, which served as the baseline for comparison. The most influential clinical predictors included clinical stage group (overall TNM staging), regional_nodes_clinical_category (N staging), tumor_clinical_category (T staging), and metastasis_clinical_category (M staging), with additional contributions from ECOG performance status, PD-L1 expression, and gender. These features consistently demonstrated high impact on the predicted hazard across patients.

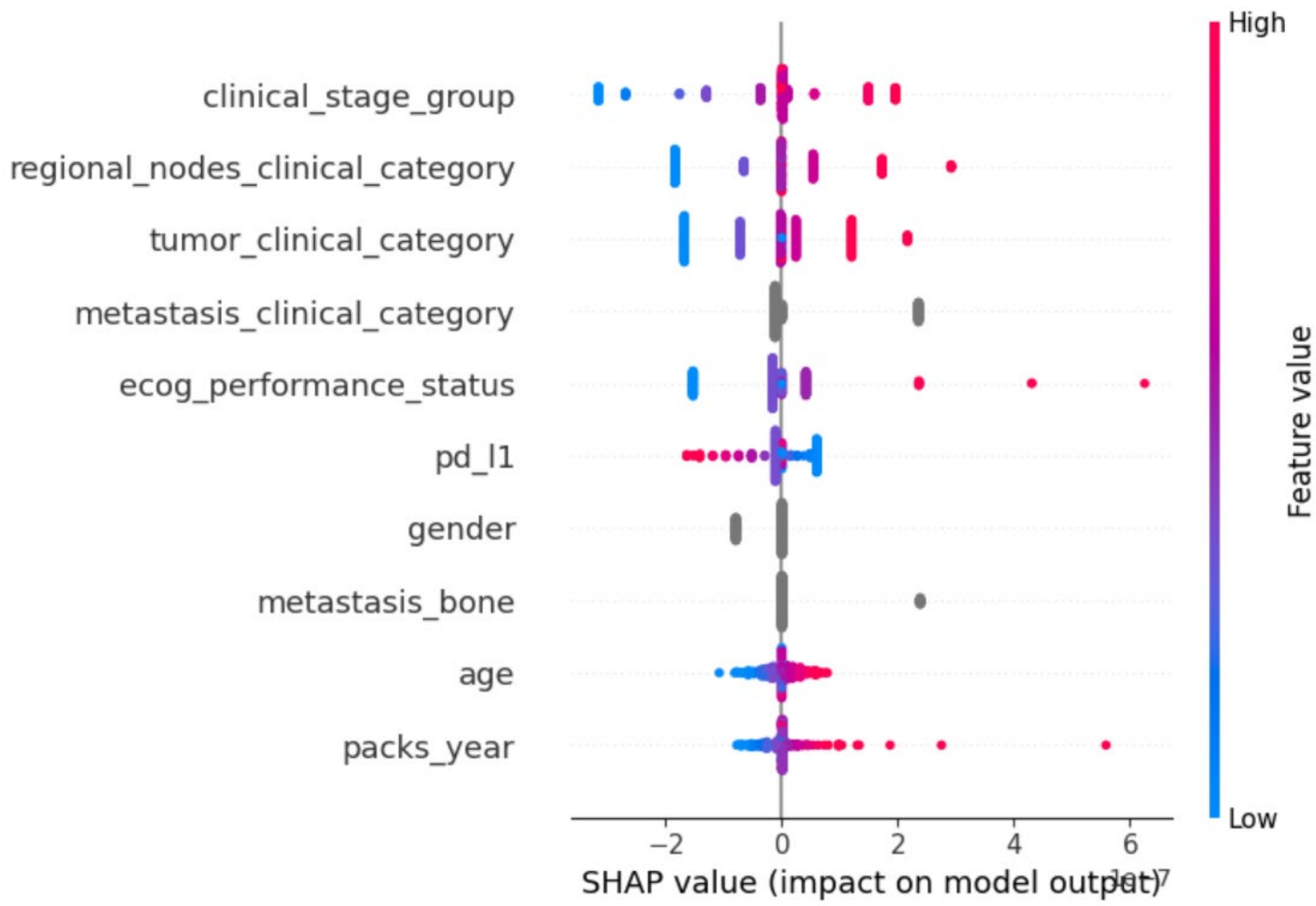

Figure 10: SHAP summary plot for the clinical-only model. The x-axis shows the SHAP value, indicating the impact of that feature on predicted survival risk. The color reflects the feature value: red for high, blue for low. Clinical stage, nodal involvement, and metastasis category showed the strongest influence on survival prediction.

Figure 11 displays the top 20 most impactful features contributing to survival prediction for the clinical + tumor texture model with ComBat harmonization. Clinical variables (e.g., metastasis category, ECOG performance status, PD-L1) and tumor texture radiomic features (e.g., GLDM, GLSZM, first-order intensity features) both contributed substantially. Radiomic features such as original_gldm_GrayLevelNonUniformity_NSCLC and original_firstorder_Skewness_NSCLC showed clear additive prognostic value alongside clinical staging variables.

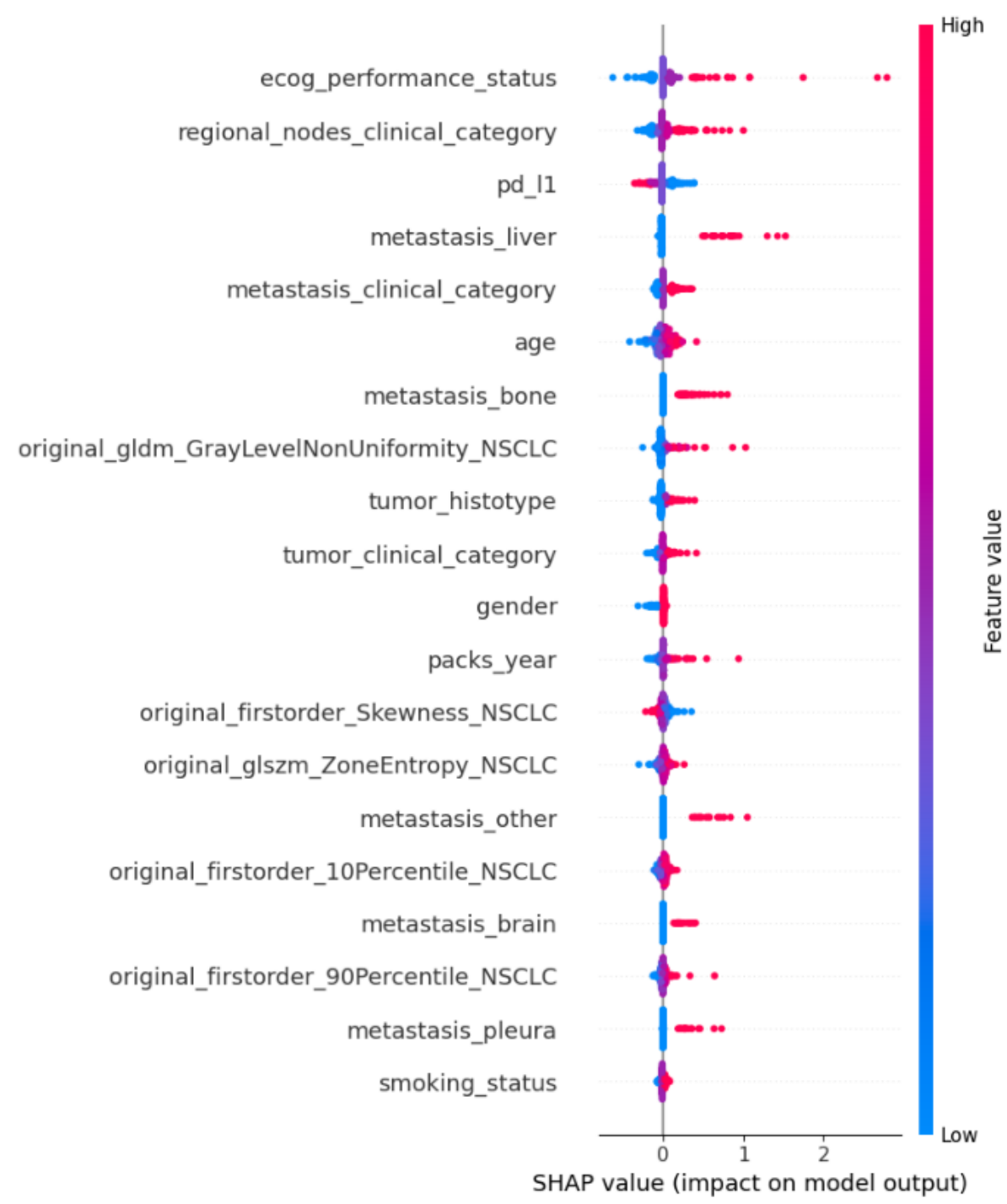

Figure 11: Depicts the SHAP summary plot for the clinical + tumor texture model with ComBat harmonization, demonstrating the notable performance among radiomics-based models. In addition to the clinical variables mentioned above, several tumor texture features (e.g., original_firstorder_Skewness, original_gldm_GrayLevelNonUniformity, original_glszm_ZoneEntropy) provided strong predictive value, emphasizing the contribution of tumor heterogeneity patterns to risk stratification.

## 3.10 Ensemble models from Combined imaging features

To explore whether combining complementary imaging features from multiple anatomical regions could enhance prognostic performance, we constructed ensemble models by averaging the predicted risk scores from selected high-performing ROI-based (only radiomic features) models. All ensemble models included ComBat-harmonized features, based on the previous results. One of the strongest performing models included tumor texture, whole lung texture, mediastinal nodes, CAC score, and FM deep features. This ensemble achieved a C-index of 0.71, 5-year t-AUC of 0.79 (95% CI: 0.70-0.87), and a hazard ratio of 3.22 (95% CI: 2.29-4.52, log rank p = 0.0000). This model captured diverse prognostic cues, integrating tumor characteristics, whole lung texture, regional spread, vascular calcification, and latent image-level features.

Other ensemble variants also showed strong performance. The combination of tumor texture + FM (cube=50) yielded a C-index of 0.71, with t-AUC of 0.79 and slightly lower HR of 3.08. Adding mediastinal nodes and CAC features to the tumor features (texture and FMCIB) further improved robustness (C-index = 0.70; t-AUC = 0.75). These results confirm that ensemble models leveraging multiple imaging domains provide consistent, clinically meaningful stratification of survival risk. Detailed performance metrics for the ensemble models are summarized in Table 4.

Table 4. Performance of ensemble models constructed from best-performing imaging feature sets (test set).
Each ensemble model was created by averaging risk scores of the test set, from selected best models. Combinations include tumor texture, mediastinal nodes, coronary artery features, whole lung texture, CAC score, and FM deep features (cube size = 50). Metrics shown are C-index, 5-year time-dependent AUC (t-AUC), hazard ratio (HR) with 95% confidence interval, and log-rank test p-values for Kaplan-Meier separation.

| **Model** | **C-index** | **Hazard ratio [CI 95%]** | **p-value (KM)** | **AUC at T=5 years** |
|---|---|---|---|---|
| Tumor (texture, ComBat)<br>MN (texture, ComBat) + MN (volume)<br>CA (texture, ComBat) + CAC score | 0.68<br>[0.63-0.72] | 3.14<br>[2.23-4.42] | 0.0000 | 0.69<br>[0.56-0.81]<br>P=0.0060 |
| Tumor (texture, ComBat)<br>FM (cube size = 50, ComBat) | 0.71<br>[0.67-0.75] | 3.08<br>[2.19-4.32] | 0.0000 | 0.79<br>[0.70-0.87]<br>P=0.0060 |
| Tumor (texture, ComBat)<br>MN (texture, ComBat) + MN (volume)<br>FM (cube size = 50, ComBat) | 0.71<br>[0.67-0.76] | 3.07<br>[2.19-4.31] | 0.0000 | 0.8<br>[0.71-0.88]<br>P=0.0000] |
| Tumor (texture, ComBat)<br>MN (texture, ComBat) + MN (volume)<br>CA (texture, ComBat) + CAC score<br>FM (cube size = 50, ComBat) | 0.70<br>[0.66-0.75] | 2.98<br>[2.13-4.18] | 0.0000 | 0.75<br>[0.63-0.86]<br>P=0.0000 |
| Whole lungs (texture, ComBat)<br>Tumor (texture, ComBat)<br>MN (texture, ComBat) + MN (volume)<br>CAC score<br>FM (cube size = 50, ComBat) | **0.71<br>[0.67-0.76]** | **3.22<br>[2.29-4.52]** | **0.0000** | **0.79<br>[0.70-0.87]<br>P=0.0000** |

## 3.11. Consensus prediction

The consensus survival model outperformed all individual ROI models, achieving an accuracy of 94.89%, sensitivity of 96.85%, specificity 70.0%, and a t-AUC of 0.9216 on the test set. Here, accuracy represents the model's ability to correctly predict a patient's binary outcome (event or non-event) at the specified time point, indicating its effectiveness in clinical prognosis. Among the individual models, the FM-based model (clinical + FM with ComBat) performed best with an t-AUC of 0.909, followed by the tumor texture model (clinical + tumor with ComBat) with t-AUC of 0.8908 and CAC score model (t-AUC = 0.8837). Full classification metrics, including sensitivity and specificity, are reported in Table 5. At this time point 5 years, 137 out of 173 valid patients (79.2%) were retained in the consensus subset, demonstrating good agreement between ROI-based models. We also evaluated the 2-year (24-month) time horizon (Supplementary Table 4). The consensus achieved t-AUC = 0.9153, with high specificity (98.5%) but lower sensitivity (62.2%), highlighting the trade-off between strict agreement and recall. The consensus subset at 2 years included 140 of 195 valid patients (71.79%). We analysed misclassifications of the consensus model at 2- and 5-years time horizons. At 5-year horizon, false negative rate (FNR) was as low as 3.15%, where only a few patients who experienced an event were misclassified as low risk. However, the false positive rate (FPR) was high at 30%, reflecting a low specificity because of the fewer true negatives in this consensus subgroup. At the 2-year horizon, FPR was 1.52% which is reflected in the low sensitivity and high FNR of 37.84% for this consensus subgroup.

Table 5. Classification performance of best-performing ROI models and their consensus at 5-year survival horizon (T = 60 months). Metrics include accuracy, sensitivity, specificity and time-dependent AUC (t-AUC).

| **Models** | **Accuracy** | **Sensitivity** | **Specificity** | **t-AUC** |
|---|---|---|---|---|
| Clinical + Lungs (texture; ComBat) | 0.8786 | 0.9266 | 0.5652 | 0.8787 |
| Clinical + Tumor (texture; ComBat) | 0.8959 | 0.9267 | 0.6957 | 0.8908 |
| Clinical + MN (texture, ComBat) | 0.8324 | 0.8333 | 0.8261 | 0.8720 |
| Clinical + CAC score | 0.8671 | 0.9133 | 0.5652 | 0.8837 |
| Clinical + FM features (cube size = 50; ComBat) | 0.9133 | 0.9467 | 0.6957 | 0.9088 |
| Consensus | 0.9489 | 0.9685 | 0.7000 | 0.9216 |

## 3.12. RQS 2.0

The methodological quality of the proposed multi-region NSCLC survival modeling framework was evaluated using the Radiomics Quality Score (RQS) 2.0 [51] framework. The study achieved a total of 30 out of 39 points, corresponding to a Radiomics Readiness Level (RRL) of 6, indicating high methodological rigor, reproducibility, and strong clinical readiness. The scoring highlighted strengths across data harmonization, multi-ROI feature integration, validation, and fairness evaluation. The cumulative progression of achieved versus maximum attainable scores per readiness level is illustrated in Figure 12, demonstrating methodological completeness for RRL-6, where criteria for calibration, explainability, and external validation were met. A detailed breakdown of all RQS 2.0 criteria with supporting evidence is provided in Appendix Table 5.

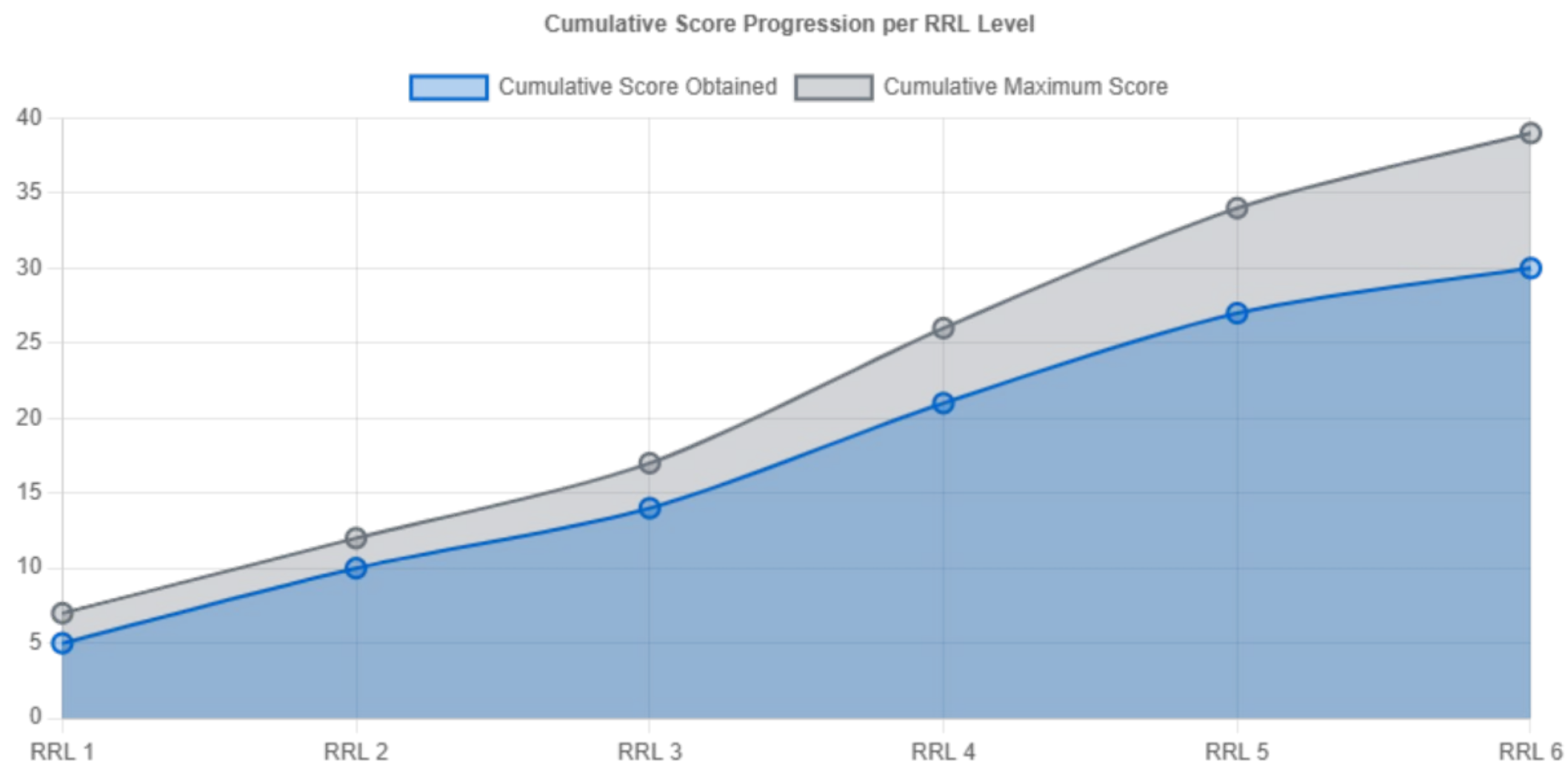

Figure 12: Cumulative Radiomics Readiness progression across RRL levels (until RRL-6).
The study achieved a RQS 2.0 score of 30/39, showing good methodological rigour and partial clinical readiness.

# 4. Discussion

In this study, we developed and systematically evaluated several prognostic survival models for non-small cell lung cancer (NSCLC) patients, using thoracic CT scans (images) and clinical data from a large multicentre cohort. Models were constructed at three levels: (1) ROI-specific handcrafted radiomics and FM deep feature radiomics models, (2) clinical + ROI combination models, and (3) harmonized versions of all the above using ComBat, RKN, and RKN+ComBat. Unlike previous studies that typically focused only on tumor-based features, we systematically analyzed texture and volumetric features from the tumor, whole lung region, mediastinal nodes (MN), coronary arteries (CA), and coronary artery calcium (CAC) scores. Features were extracted from both handcrafted radiomic features and pretrained FM deep features derived from 3D image patches at multiple scales. Survival prediction was performed using regularized Cox proportional hazards models, optimized in cross-validation. Evaluation metrics included C-index, 5-year time-dependent AUC, and hazard ratios from Kaplan-Meier stratification. Feature importance was interpreted using SHAP (SHapley Additive exPlanations) analysis.

The clinical + FMCIB model (cube size = 50 and 128) harmonized with ComBat achieved one of the strongest performance with a C-index = 0.76, t-AUC = 0.89, and HR range: 4.75-5.01. This validates the usefulness of deep features generated on 3D patches through pretrained foundation models especially when they are harmonised and integrated with clinical information. The clinical + tumor texture model showed a high level of prognostic performance also (C-index = 0.76; t-AUC = 0.88), which indicated the complementary value of handcrafted radiomic features. As indicated in the calibration analysis (Appendix Figure 1) the ComBat-harmonized tumor-texture model aligns well to the observed 5-year survival probabilities, which indicates that the overall calibration is good and the model is able to effectively estimate risk across the cohort. These results indicate that although deep learning-learned features provide rich and hierarchical representations, conventional radiomics may still encode important prognostic information especially when harmonized through domain-adapted pipelines. Beyond tumor-centric analysis, our results highlight the independent prognostic contribution of additional ROIs. Whole lung texture features capture global parenchymal changes that might be linked to comorbidities in lung cancer patients such as fibrosis or emphysema [52,53]. MN and CA Radiomic models also gave prognostic signals, which models regional spread and cardiovascular burden. Alone, CAC scores are weak predictors, but, when added to clinical variables, provide prognostic information. Their inclusion in ensemble models enhanced overall performance, which represents their contribution to cardiovascular burden, which is a known prognostic factor in cancer populations [54]. External validation on an independent open source dataset (NSCLC-Radiomics [6,41]) also confirmed the added value of multi anatomical regions in NSCLC prognosis (Appendix Table 2)

FM deep features, extracted from 3D image patches using a pretrained foundation model [31], achieved comparable performance to handcrafted radiomics-based models without the need of radiomic features. The 50 voxel cube size produced the highest performance (C-index = 0.76; t-AUC = 0.88), which was higher than 128 and 96 patch sizes used in FM deep feature models. This finding is in line with the fact that the foundation model was initially trained on 50 voxel patches, so it could be best suited to find meaningful features on inputs of equal size. Even though the 128 cube more accurately reflected the sizes of the tumors in our data (see Appendix Table 1), it may have included too much surrounding tissue, reducing the focus on the tumor itself. Conversely, the 50 patch size probably focused on the core lesion and therefore the prognostic features were stronger and more reliable.

When multiple ROI features were combined into ensemble models using soft-voting (i.e., averaging risk scores), performance improved further. The best-performing ensemble, which integrated ComBat-harmonized features from the tumor, lungs, mediastinal nodes, CAC score, and FM deep features, achieved a C-index of 0.71 and a t-AUC of 0.79. These results demonstrate the additive value of multi-region imaging features in survival stratification.

To complement time-to-event modeling, we derived binary classifications at clinically relevant survival horizons by thresholding the predicted survival probability $S(t)$ from each model using Youden's index. We then implemented a strict consensus strategy [55–58] across the best-performing ROI models, retaining predictions only for patients where all models agreed on the binary outcome. This high-confidence subset demonstrated robust predictive performance: at the 5-year horizon, the consensus model achieved a t-AUC of 0.92, sensitivity of 96.9%, and specificity of 70.0%, while covering 79% of valid patients. At 2 years, consensus maintained a strong t-AUC of 0.9153, with high specificity (98.5%) but reduced sensitivity (62.16%). The failure model analysis at the two time horizons show that at earlier time points the model may be more conservative to flag patients and may run a risk of missing a higher proportion of those who eventually experienced an event. The findings highlight that consensus model based predictions may show trade-offs across time points favouring sensitivity at longer horizons and specificity at shorter horizons. Overall, these results highlight the potential of consensus modeling for prioritizing actionable risk predictions across heterogeneous feature sets, particularly in multi-ROI contexts.

An important methodological insight of our work is image-level and feature-level harmonization when the data under observation is multicentric. We individually applied RKN to the whole lung region and ComBat to all the extracted features, while also integrating them together to observe if they act synergistically or competitively. Reconstruction-kernel normalization (RKN) [25] first attenuates high-frequency differences introduced by sharp versus soft CT kernels, bringing texture appearance closer to a common reference. A subsequent ComBat [25,50] correction is then applied to the extracted features, shrinking residual centre-specific means and variances while preserving biological signal. This cascaded approach, applying RKN followed by ComBat, was particularly effective for tumor texture features, boosting 5-year t-AUC from 0.73 (no harmonization) to 0.75 with ComBat alone, and further to 0.76 with combined RKN+ComBat harmonization. Notably, this synergistic benefit was observed for several regions beyond the tumor. Lung texture models also showed consistent, though smaller, performance gains when both RKN and ComBat were applied sequentially. These findings underscore that correcting both low-level image differences and high-level feature distributions may help achieve optimal cross-site generalizability in CT-based survival models. Moreover, through our results, we demonstrate that foundation-model (FM) embeddings are not inherently centre-agnostic and may still suffer from data heterogeneity unless systematically harmonised. Although FM deep features are often assumed to be robust to technical variability due to their unsupervised large-scale pretraining , our results show otherwise. Single-pass ComBat harmonization improved the performance of FM features extracted from 50 cube voxel patches, raising the C-index from 0.66 to 0.67 and t-AUC from 0.66 to 0.74. In contrast, FM embeddings extracted from 96 cube size performed poorly (C-index 0.51) even after harmonization, highlighting that the choice of patch size and the application of batch correction must be carefully tuned together for optimal survival prediction. These observations are highly relevant given that most previous multi-centre radiomics studies have evaluated either RKN or ComBat independently and rarely assessed their combined application. Furthermore, prior works focused almost exclusively on handcrafted features, with little attention paid to harmonization strategies for foundation-model-derived deep features. Our results therefore contribute by addressing an important gap, offering a practical template for harmonization pipelines that can be generalized across both traditional radiomics and modern FM-based approaches in real-world heterogeneous clinical networks.

Harmonization remains a critical requirement for radiomics and deep-features-based modeling especially in multi-centre settings where variations in scanner hardware, reconstructions settings, and imaging protocols introduce significant technical biases. In our prior review [19], we outlined how unaddressed acquisition variability can inflate false associations, reduce generalizability, and compromise model reproducibility across sites. As multi-institutional imaging repositories grow, reliance on harmonization strategies will become even more essential for ensuring robust, clinically deployable models. Our study uniquely illustrates that both image-domain harmonisation (RKN) and feature-domain harmonization (ComBat) can be applied to maximize correction effectiveness, across both traditional radiomic features and FM free features. Furthermore, our findings show that even features from pretrained FMs, often presumed to be robust, are susceptible to acquisition biases unless appropriate harmonization steps are integrated. Thus, addressing harmonization systematically, across imaging and feature domains, is not merely an auxiliary step but a foundational prerequisite for achieving reproducibility, fairness and cross-site clinical translation of radiomics and deep imaging biomarkers. We applied ComBat harmonization using centre as the batch variable, as centre-level differences often encapsulate scanner and protocol variability, and ensure sufficient sample sizes for stable parameter estimations. While some centres operated multiple scanners, scanner-level harmonization was not pursued due to limited batch sizes and potential metadata inconsistencies, though future work could explore this granularity.

Previous studies have explored the integration of radiomic and clinical features for survival prediction in NSCLC. Hou et al. [59] developed a deep learning model combining radiomic and clinical features, achieving C-index values of 0.74 to 0.75 at 8, 12, and 24 months post-diagnosis. Braghetto et al. [60] evaluated radiomics and deep learning-based approaches on the LUNG1 dataset, reporting improvements in AUC values when combining radiomic and deep features. However, these studies primarily focused on tumor regions and did not comprehensively assess multiple ROIs or incorporate FM deep features. Ferretti et al. [61] proposed a 3D convolutional autoencoder trained from scratch to extract deep features from tumor volumes, which, when combined with radiomic and clinical features, improved survival prediction. Their multi-domain signature achieved a C-index of 0.6309. While their approach focused on tumor-centric features, our study extends this by incorporating multiple ROIs and utilizing FM deep features extracted from a pretrained model, thereby enhancing the comprehensiveness and potential generalizability of the prognostic models.

While this study provides valuable insights into survival prediction for lung cancer patients, several limitations should be acknowledged. Firstly, the retrospective design and reliance on pre-existing datasets may introduce selection bias. The generalizability of the models to other populations, imaging protocols, especially outside the platform, requires further validation. Secondly, the traditional calculation of the Agatston score, which multiplies the area of calcified plaque by a density weighting factor, assumes that both higher volume and higher density of CAC are associated with increased cardiovascular risk. However, Criqui et al. [62] demonstrated that, at any given CAC volume, higher CAC density was inversely associated with the risk of coronary heart disease and cardiovascular disease, while CAC volume was positively associated with risk. This finding suggests that the conventional Agatston scoring method may not fully capture the nuanced relationship between CAC characteristics and cardiovascular risk, potentially leading to misclassification in risk stratification.

Future research should focus on prospective studies to assess the clinical utility of these models in real-world settings. Integrating additional data modalities, such as genomic and histopathological information, could provide a more comprehensive understanding of tumor biology and patient prognosis. Moreover, refining CAC scoring methods to account for both volume and density may enhance the accuracy of cardiovascular risk assessment in NSCLC patients.

# 5. Conclusion

This study demonstrates that combining harmonized, both at the image-level and feature-level domains, region-specific radiomics and foundation model deep features with clinical data can enable robust, interpretable, and generalizable survival prediction in non-small cell lung cancer (NSCLC) using routine thoracic CT. By systematically evaluating models across tumor, lung, mediastinal nodes, coronary arteries, and coronary artery calcium (CAC), and applying harmonization techniques such as ComBat and RKN, multi-centre variability can be effectively addressed to improve model reliability. The proposed pipeline, integrating both handcrafted radiomic features and pretrained foundation model embeddings, achieved strong prognostic performance, with concordance index values up to 0.76 and five-year survival time-dependent AUCs reaching 0.89. Ensemble approaches further enhanced the performance of imaging-based models.

In addition, consensus analysis across the best-performing region-specific models identified a high-confidence subset of patients for whom all models agreed on the binary outcome. This subset covered up to 79 percent of the cohort and achieved the highest five-year time-dependent AUC observed (0.922), along with excellent sensitivity (96.9 percent). These findings indicate that model agreement across diverse anatomical regions is associated with more reliable prognostic signals. Overall, our results support the clinical potential of harmonized CT-derived imaging feature, across both traditional radiomics and foundation model representation, for individualized risk stratification and enhanced interpretability in multicentre lung cancer survival modeling.

# Code

The codes and data analysis scripts are available on Github repository
https://github.com/shruti26mali/PixelsToPrognosis-NSCLC

# Grants and funding

Authors acknowledge financial support from ERC advanced grant (ERC-ADG-2015 n° 694812 - Hypoximmuno), ERC-2020-PoC: 957565-AUTO.DISTINCT. Authors also acknowledge financial support from the European Union's Horizon research and innovation programme under grant agreement: CHAIMELEON n° 952172 (main contributor), ImmunoSABR n° 733008, EuCanImage n° 952103, TRANSCAN Joint Transnational Call 2016 (JTC2016 CLEARLY n° UM 2017-8295), IMI-OPTIMA n° 101034347, AIDAVA (HORIZON-HLTH-2021-TOOL-06) n°101057062, REALM (HORIZON-HLTH-2022-TOOL-11) n° 101095435, RADIOVAL (HORIZON-HLTH-2021-DISEASE-04-04) n°101057699 and EUCAIM (DIGITAL-2022-CLOUD-AI-02) n°101100633.

# Disclosures:

Disclosures from the last 36 months within and outside the submitted work: none related to the current manuscript; outside of current manuscript: grants/sponsored research agreements from Radiomics SA, Convert Pharmaceuticals and LivingMed Biotech. He received a presenter fee (in cash or in kind) and/or reimbursement of travel costs/consultancy fee (in cash or in kind) from Radiomics SA, BHV & Roche. PL has shares in the companies Radiomics SA, Convert pharmaceuticals, Comunicare, LivingMed Biotech, BHV and Bactam. PL is co-inventor of two issued patents with royalties on radiomics (PCT/NL2014/050248 and PCT/NL2014/050728), licensed to Radiomics SA; one issued patent on mtDNA (PCT/EP2014/059089), licensed to ptTheragnostic/DNAmito; one non-issued patent on LSRT (PCT/P126537PC00, US: 17802766), licensed to Varian; three non-patented inventions (softwares) licensed to ptTheragnostic/DNAmito, Radiomics SA and Health Innovation Ventures and two non-issued, non-licensed patents on Deep Learning-Radiomics (N2024482, N2024889). He confirms that none of the above entities were involved in the preparation of this paper.

# Appendix

Table 1. Voxel dimensions of the largest tumor per patient in the training set. Descriptive statistics and custom quantiles for tumor size along each axis. The 95th percentile values support the selection of a 128×128×128 patch size for FM deep feature extraction.

| Statistic / Quantile | x_dim | y_dim | z_dim |
|---|---|---|---|
| **Mean** | 46.01 | 47.08 | 50.67 |
| **Standard deviation** | 28.24 | 29.21 | 38.78 |
| **Min** | 3 | 3 | 2 |
| **25th percentile** | 25.00 | 24.25 | 24.00 |
| **Median (50%)** | 40.00 | 41.50 | 42.00 |
| **75th percentile** | 63.00 | 63.00 | 65.75 |
| **95th percentile** | 101.70 | 101.00 | 129.35 |
| **Max** | 137.00 | 169.00 | 249.00 |

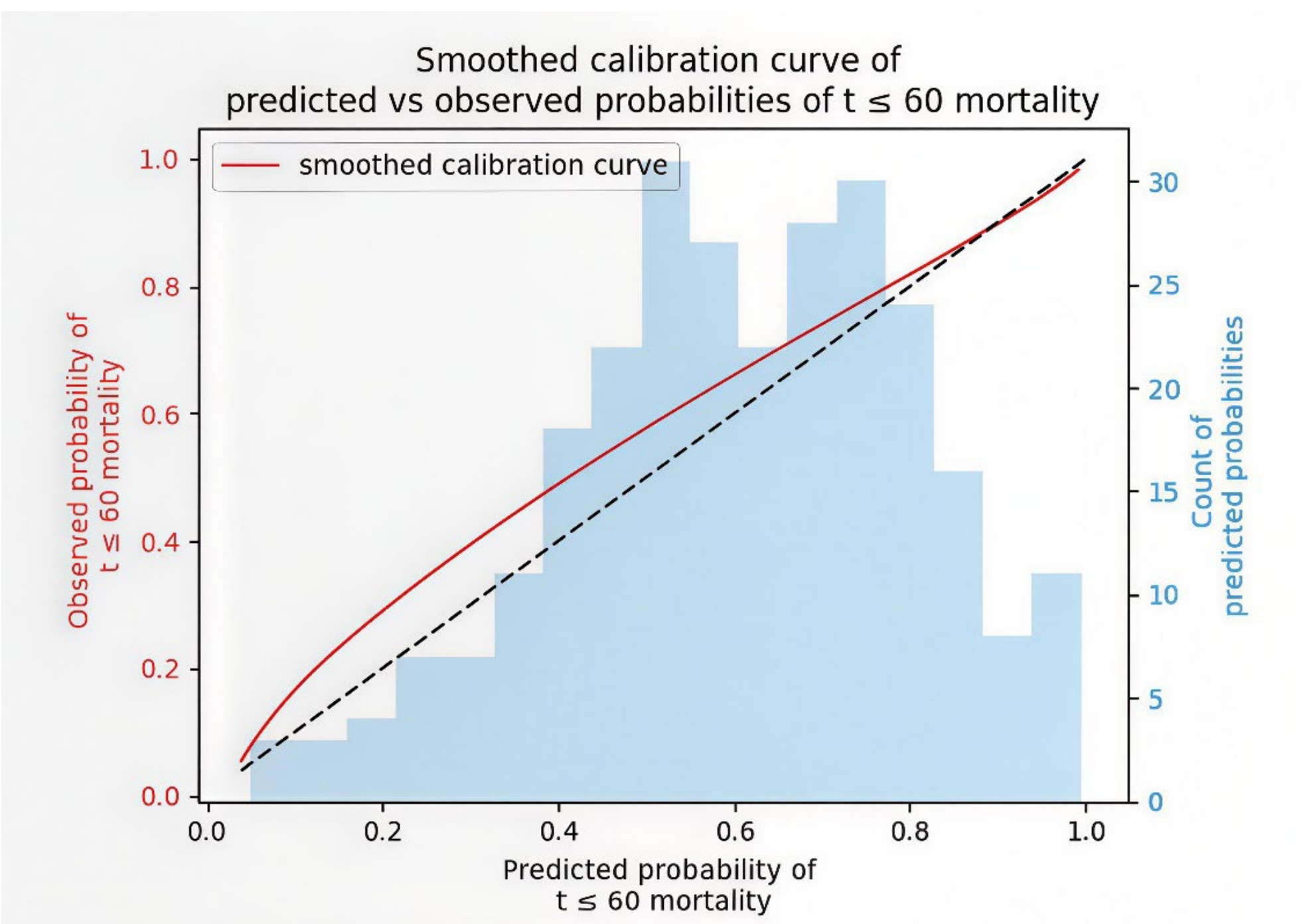

Figure 1. Calibration curve for the radiomics model (tumor texture with ComBat harmonization).
The plot shows the relationship between the predicted probability of 5-year mortality (x-axis) and the observed probability of mortality (y-axis). The red smoothed curve represents the model's calibration, while the dashed black line indicates the ideal reference (perfect calibration). The histogram (blue) displays the distribution of predicted probabilities across the cohort. The curve demonstrates good overall agreement between predicted and observed survival probabilities, with slight underestimation of mortality risk at lower probabilities and near-perfect alignment toward higher risk estimates, indicating reliable model calibration.

**External validation on the NSCLC Radiomics (LUNG1) dataset**

To assess the generalizability of the proposed radiomic models, external validation was performed using the open-access **NSCLC Radiomics (LUNG1)** dataset. The trained models from our multicentre CHAIMELEON cohort were directly applied to this dataset without retraining or fine-tuning. Radiomic features were standardized using the same preprocessing and scaling parameters as in the internal test set.

The external validation results (Table 2) show that the models achieved moderate prognostic performance, with C-index values ranging between 0.50 and 0.59 and 5-year t-AUC values between 0.51 and 0.60. The tumor volume model demonstrated the highest concordance (C-index = 0.59, HR = 1.32 [1.10–1.63], p = 0.0064), followed by the CAC score model (C-index = 0.53, HR = 1.26 [1.02–1.54], p = 0.0289). Although lower than internal validation results, these findings confirm that the handcrafted radiomics features retain measurable prognostic signals across independent datasets.

Table 2. External validation results on NSCLC Radiomics (LUNG1) dataset (radiomic features from tumor, mediastinal nodes, coronary arteries, and CAC scores).

| Model | Test<br>(C-index) | Test<br>(Hazard ratio [CI 95%])) | p-value<br>(KM) | Test<br>(AUC at T=5 yrs)<br>[CI 95%]<br>P-value |
|---|---|---|---|---|
| Tumor (volume) | 0.59<br>[0.55 -0.62] | 1.32<br>[1.1-1.63] | 0.0064 | 0.57<br>[0.5 -0.64]<br>P=0.0600 |
| Tumor (texture) | 0.57<br>[0.53 -0.6] | 1.2<br>[1.0-1.47] | 0.0862 | 0.52<br>[0.45 -0.59]<br>P=0.5200 |
| MN (volume) | 0.55<br>[0.52 -0.58] | 1.14<br>[0.92-1.41] | 0.2387 | 0.60<br>[0.53 -0.67]<br>P=0.0020 |
| MN (texture) | 0.51<br>[0.48 -0.55] | 1.0<br>[0.79-1.21] | 0.8284 | 0.58<br>[0.50 -0.65]<br>P=0.0460 |
| CA (texture) | 0.5<br>[0.47 -0.53] | 1.08<br>[0.88-1.33] | 0.4487 | 0.55<br>[0.48 -0.62]<br>P=0.1580 |
| CAC score | 0.53<br>[0.50 -0.57] | 1.26<br>[1.02-1.54] | 0.0289 | 0.51<br>[0.49 -0.53]<br>P=0.1640 |

**Subgroup analysis using unharmonized tumor texture model:**

To assess how acquisition parameters and different centres affect model reproducibility, one of the best handcrafted radiomics models (tumor texture, unharmonized) was evaluated separately across imaging centres and scanner manufacturers. Results showed moderate fluctuations in prognostic performance (C-index = 0.49-0.85; HR = 1.3-7.4; t-AUC = 0.49-0.85), indicating heterogeneity related to site- and scanner-specific factors.

Table 3: Subgroup analysis using the unharmonized tumor texture model.
The table reports the C-index, hazard ratio (95 % CI), log-rank p-values, and 5-year time-dependent AUC across individual centres and scanner manufacturers.

| Model | Test<br>(C-index) | Test<br>(Hazard ratio [CI 95%])) | p-value<br>(KM)<br>TEST | Test<br>(AUC at T=5 yrs)<br>[CI 95%]<br>P-value |
|---|---|---|---|---|
| Centre 1 (LaFe)<br>(nTs=111) | 0.66<br>[0.60 -0.71] | 1.93<br>[1.26-2.96] | 0.0027 | 0.77<br>[0.63 -0.91]<br>P=0.0000 |
| Centre 3 (ULS)<br>(nTs=44) | 0.67<br>[0.19 -0.91] | 1.42<br>[0.19-10.56] | 0.7313 | NA (max follow-up < 5 years) |
| Centre 6 (CHU Angers)<br>(nTs=81) | 0.69<br>[0.60 -0.77] | 2.07<br>[1.10-3.89] | 0.0235 | 0.80<br>[0.60 -0.98]<br>P=0.0100 |
| Centre 8 (CHU Nimes)<br>(nTs=30) | 0.49<br>[0.21 -0.77] | 1.32<br>[0.50-3.49] | 0.5694 | 0.49<br>[0.21 -0.77]<br>P=1.0680 |
| Centre 9 (Paris St-Joseph)<br>(nTs=6) | 0.49<br>[0.00 -0.89] | 1.32<br>[0.18-9.53] | 0.7863 | NA (max follow-up < 5 years) |
| GE MEDICAL SYSTEMS<br>(nTs=67) | 0.60<br>[0.50 -0.70] | 1.83<br>[0.94-3.54] | 0.0736 | 0.77<br>[0.55 -0.94]<br>P=0.0220 |
| Philips<br>(nTs=123) | 0.66<br>[0.60 -0.72] | 2.10<br>[1.33-3.33] | 0.0015 | 0.70<br>[0.52 -0.87]<br>P=0.0320 |
| Siemens<br>(nTs=56) | 0.64<br>[0.49 -0.78] | 2.05<br>[0.89-4.76] | 0.0934 | 0.62<br>[0.36 -0.85]<br>P=0.3691 |
| TOSHIBA<br>(nTs=23) | 0.85<br>[0.62 -1.00] | 7.44<br>[1.96-28.22] | 0.0032 | 0.85<br>[0.62 -1.00]<br>P=0.0080 |
| MALE<br>(nTs=178) | 0.68<br>[0.62 -0.73] | 2.20<br>[1.49-3.26] | 0.0001 | 0.73<br>[0.61 -0.83]<br>P=0.0000 |
| FEMALE<br>(nTs=94) | 0.66<br>[0.58 -0.74] | 2.53<br>[1.40-4.56] | 0.0021 | 0.71<br>[0.47 -0.90]<br>P=0.0980 |
| AGE < median age (66 years)<br>(nTs=134) | 0.62<br>[0.55 -0.69] | 1.89<br>[1.17-3.07] | 0.0099 | 0.70<br>[0.53 -0.84]<br>P=0.0240s |
| AGE > median age (66 years)<br>(nTs=138) | 0.71<br>[0.66 -0.76] | 3.27<br>[2.09-5.13] | 0.0000 | 0.78<br>[0.66 -0.89]<br>P=0.0000 |

Table 4: Consensus models for survival at T = 2 years :
Valid cases (survival cases at T) = 195/272
Consensus coverage = 140 / 195 (71.79%)

| Models | Accuracy | Sensitivity | Specificity | t-AUC |
|---|---|---|---|---|
| Clinical + Lungs (texture; ComBat) | 0.7128 | 0.5478 | 0.9500 | 0.8694 |
| Clinical + Tumor (texture; ComBat) | 0.7128 | 0.5565 | 0.9375 | 0.8730 |
| Clinical + MN (texture, ComBat) | 0.7385 | 0.6087 | 0.9250 | 0.8773 |
| Clinical + CAC score | 0.7333 | 0.5826 | 0.9500 | 0.8666 |
| Clinical + FM features<br>(cube size = 50; ComBat) | 0.7333 | 0.6174 | 0.9000 | 0.8909 |
| Consensus | 0.7929 | 0.6216 | 0.9849 | 0.9153 |

Supplementary Table 5. Detailed Radiomics Quality Score (RQS 2.0) evaluation for the proposed NSCLC prognosis framework. The study achieved an RQS 2.0 score of 30/39 (Radiomics Readiness Level 6).

| No. | Criteria | Selected Option | Points | Explanation |
|---|---|---|---|---|
| **RRL 1 - Foundational Exploration** | | | | |
| 1 | Unmet Clinical Need – Unmet clinical need (UCN) defined.<br>● UCN is agreed upon and defined by more than one centre.<br>● UCN is defined using an established consensus method such as the Delphi method. | Implemented: Delphi method (+2) | 2 | UCN defined and endorsed via consensus across 5 CHAIMELEON centers showing multi-centre agreement on UCN in lung cancer use case |
| 2 | Hardware Description – Detailed description of the imaging hardware used, including model, manufacturer, and technical specifications. | Implemented (+1) | 1 | scanner manufacturer & model reported (refer methods sections) |
| 3 | Image Protocol Quality – Five levels of image protocol quality for TRIAC:<br>● Level 0: Protocol not formally approved.<br>● Level 1: Approved with a reference number in the institutional archive.<br>● Level 2: Approved with formal quality assurance (recommended minimum for prospective trials).<br>● Level 3: Established internationally; published in guidelines and peer-reviewed papers.<br>● Level 4: Future proof (follows TRIAC Level 3, FAIR principles, retains raw data). | Not implemented | 0 | No formal or standardized imaging protocol across centers; institutional approval documentation not available |
| 4 | Inclusion and Exclusion Criteria – Detailed criteria for patient selection in studies, including rationale. | Implemented (+1) | 1 | clear criteria given in Methods |
| 5 | Diversity and Distribution – Identify potential biases before the project (demographics, socioeconomic, geographic, medical profiles). | Implemented (+1) | 1 | Patient demographic and acquisition heterogeneity were reported (see Method and Results section) |
| **RRL 2 - Data Preparation** | | | | |
| 6 | Feature Robustness – Assess robustness via:<br>1. Imaging at multiple time points (test–retest).<br>2. Multiple segmentations (different physicians/algorithms/noise/perturbations).<br>3. Phantom study (identify inter-scanner/vendor differences). | Implemented (+1) | 1 | Robustness against test–retest or inter-observer variation was not evaluated; however, scanner- and centre-wise performance of the unharmonized model was analyzed (appendix). |
| 7 | Preprocessing of Images – Apply steps to standardize images with clear reasoning. | Implemented (+1) | 1 | Image voxel resampling done prior to feature extraction, refer to Methods section |
| 8 | Harmonization – Use image-level (e.g. CycleGANs) or feature-level (e.g. ComBat) harmonization techniques. | Implemented (+1) | 1 | Both image-level (RKN) and feature-level (ComBat) harmonization were applied, and their combination evaluated to reduce acquisition variability. |
| 8 | Compliance with International Standards – Use implementations that adhere to standards (e.g., IBSI) for radiomic feature extraction. | Implemented (+1) | 1 | All handcrafted features were extracted using an IBSI-compliant tool (pyradiomics) ensuring reproducibility and standardization. |
| 10 | Automatic Segmentation – Use an automated segmentation algorithm for ROI definition. | Implemented (+1) | 1 | ROIs including lung, tumor, mediastinal nodes, and coronary arteries were automatically segmented |
| **RRL 3 - Prototype Model Development** | | | | |
| 11 | Feature Reduction – Reduce features to lower the risk of overfitting (especially when features outnumber samples; check for correlations with volume). | Implemented (+1) | 1 | Feature reduction was performed using correlation filtering and Optuna-based model optimization to avoid multicollinearity and improve model generalization. |
| 12 | Feature Robustness for Feature Selection – Integrate robustness evaluation into feature selection using prior test–retest, phantom, or segmentation studies. | Not implemented | 0 | No dedicated test–retest, phantom, or inter-observer robustness filtering was used during feature selection. |
| 13 | HCR + DL Combination – Compare and explore the synergistic combination of handcrafted radiomics and deep learning models. | Implemented (+1) | 1 | Ensemble models combined handcrafted radiomics (HCR) with FM-derived deep features, showing complementary prognostic contributions. |
| 14 | Multivariable Analysis – Incorporate non-radiomics features (clinical, genomic, proteomic) to yield a holistic model. | Implemented (+2) | 2 | Clinical variables were combined with radiomic and FM deep features to develop comprehensive prognostic models that improved C-index and HR performance. |
| **RRL 4 - Internal Validation** | | | | |
| 15 | Single Center Validation – Validation performed on data from the same institute without retraining or adapting the cut-off value. | Implemented (+1) | 1 | Internal validation was conducted within the multicentre dataset (see centre-wise results in appendix) |
| 16 | Cut-off Analyses – Identify optimal thresholds (e.g., using Youden's Index) for classification or survival analysis. | Implemented (+1) | 1 | Youden's Index was applied to define optimal thresholds for binary classification from survival probabilities in the consensus experiment. |
| 17 | Discrimination Statistics – Report discrimination metrics (e.g., ROC curve, sensitivity, specificity) with significance (p-values, CIs).<br>● Statistic reported<br>● With Resampling method | Resampling method applied (+2) | 2 | Discrimination metrics (C-index, time-dependent AUC, HR, p-values) were reported using bootstrapped confidence intervals and cross-validation. |
| 18 | Calibration Statistics – Report calibration metrics (e.g., calibration-in-the-large, slope, plots). | Implemented (+1) | 1 | A calibration curve was plotted for the best-performing HRF model (tumor texture + ComBat) to assess alignment between predicted and observed survival probabilities. |
| 19 | Failure Mode Analysis – Document model limitations with examples of edge cases. | Implemented (+1) | 1 | Failure mode analysis was performed for the consensus model, identifying the distribution of false positives and false negatives (results section). |
| 20 | Open Science and Data – Make code and data publicly available.<br>● Open scans (+1)<br>● Open segmentations (+1)<br>● Open code (+1) | One aspect (+1) | 1 | The preprocessing code are available via open repositories on github |
| **RRL 5 - Capability Testing** | | | | |
| 21 | Multi-centre Validation – Validation with data from multiple institutes ensuring no overlap:<br>● One external institute<br>● Two or more external institutes<br>● Third-party platform with completely unseen data | One institute (+1) | 1 | External validation was performed using the publicly available LUNG1 (NSCLC Radiomics) dataset, serving as an independent external cohort (appendix). |
| 22 | Comparison with 'Current Clinical Standard' – Assess model agreement or superiority versus the current gold standard (e.g., TNM staging). | Implemented (+2) | 2 | Model outputs for clinical TNM staging only model was reported |
| 23 | Comparison to Previous Work – Compare performance with published HCR signatures or DL algorithms. | Implemented (+1) | 1 | Results were benchmarked against prior studies (refer discussion section) |
| 24 | Potential Clinical Utility – Report on the current and potential clinical application (e.g., decision curve analysis). | Implemented (+2) | 2 | The consensus and survival models stratified patients into distinct risk groups with significant survival differences, supporting |
| | | | | their potential clinical utility. More content in the discussions section. |
| | **RRL 6 - Trustworthiness Assessment** | | | |
| 25 | Explainability – Apply explainability tools (e.g., SHAP for HCR, GradCAM for DL) to clarify model predictions. | Implemented (+1) | 1 | Shap analysis plots provided to show clinical and radiomics predictors (results section) |
| 26 | Explainability Evaluation – Conduct qualitative and quantitative evaluations of interpretability methods (e.g., checking consistency to adversarial perturbations). | Not implemented | 0 | No explainability evaluation carried out |
| 27 | Biological Correlates – Detect and discuss biological correlates to deepen understanding of radiomics and underlying biology. | Implemented (+1) | 1 | Discussed biological relevance: e.g., texture features linked to tumor heterogeneity, whole-lung parenchymal changes (fibrosis/emphysema), and CAC reflecting cardiovascular burden |
| 28 | Fairness Evaluation and Mitigation – Evaluate model performance for biases and apply bias correction if needed.<br>● Fairness evaluated<br>● Bias correction applied | Fairness evaluated (+1) | 1 | No bias correction applied but fairness was evaluated for different subgroups (age/sex) in the appendix. |
| | Total = 30/39 (77%) | | | |